\documentclass{article}

\usepackage{arxiv}
\usepackage[utf8]{inputenc} 
\usepackage[T1]{fontenc}    
\usepackage{hyperref}       
\usepackage{url}            
\usepackage{booktabs}       
\usepackage{nicefrac}       
\usepackage{microtype}      
\usepackage{lipsum}
\usepackage{multirow}
\usepackage{siunitx}
\usepackage{url}
\usepackage{bm}
\usepackage{hyperref}
\usepackage{amsmath,amssymb,amsfonts}

\usepackage{algorithm,algpseudocode}
\usepackage{graphicx, tikz}
\usepackage{natbib} 

\usepackage{tabularx}
\usepackage{makecell}
\usepackage{setspace}
\usepackage{adjustbox}
\usepackage{makecell}
\usepackage{xspace}
\usepackage{microtype}
\usepackage{subfigure}
\usepackage{fancyhdr}
\usepackage{framed}
\usepackage{rotating}

\title{A Review on Visual Privacy Preservation Techniques for Active and Assisted Living}

\author{\href{https://orcid.org/0000-0002-2301-569X}{\includegraphics[scale=0.06]{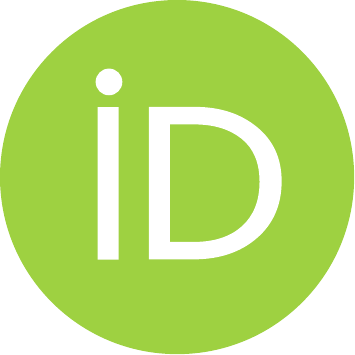}\hspace{1mm}Siddharth Ravi\thanks{Corresponding author. © 2021. This manuscript version is made available under the CC-BY-NC-ND 4.0 license \href{https://creativecommons.org/licenses/by-nc-nd/4.0/}{https://creativecommons.org/licenses/by-nc-nd/4.0/}}} \\
	Department of Computing Technology\\
	University of Alicante, \\
	Alicante, 03690, Valencian Community, Spain \\
	\texttt{siddharth.ravi@ua.es} \\
	\And
	Pau Climent-Pérez \\
	Department of Computing Technology\\
	University of Alicante, \\
	Alicante, 03690, Valencian Community, Spain \\
	\texttt{pcliment@dtic.ua.es}
	\And
	Francisco Florez-Revuelta \\
	Department of Computing Technology\\
	University of Alicante, \\
	Alicante, 03690, Valencian Community, Spain \\
	\texttt{francisco.florez@gcloud.ua.es} \\
}

\begin{document}

\maketitle

\begin{abstract}
	This paper reviews the state of the art in visual privacy protection techniques, with particular attention paid to techniques applicable to the field of active and assisted living (AAL). A novel taxonomy with which state-of-the-art visual privacy protection methods can be classified is introduced. Perceptual obfuscation methods, a category in the taxonomy, is highlighted. These are a category of visual privacy preservation techniques particularly relevant when considering scenarios that come under video-based AAL monitoring. Obfuscation against machine learning models is also explored. A high-level classification scheme of the different levels of privacy by design is connected to the proposed taxonomy of visual privacy preservation techniques. Finally, we note open questions that exist in the field and introduce the reader to some exciting avenues for future research in the area of visual privacy.
\end{abstract}


\section{Introduction}
\label{sec:intro}
There are two major reasons for visual privacy to be preserved. One is identity protection, where the identity of the person in a visual is to be hidden from entities who might analyse the feed without the necessary access privileges. Following convention, these entities will be addressed  as \textit{adversaries} in this review. Adversaries can either be machine learning models that train on data collected without user consent, or persons who view sensitive visuals without being provided with the necessary consent. The second reason for visual privacy is the preservation of trust for persons who require monitoring. 
 
Active and Assisted Living (AAL) is considered as the primary area of interest while choosing the methods surveyed. In active and assisted living scenarios, computer vision systems provide support to the older section of society, assisting them in their everyday tasks within a care home or a nursing home, a living community, or in their private homes. A typical AAL care home might be equipped with RGB cameras, the feeds of which are monitored and analysed to provide support to the residents in time of need. In these cases, the identity of the resident is of relatively less interest, as that is usually of a more public nature. A typical AAL care home resident, for example, could have given consent for them to be monitored by the home's personnel and their family for safety reasons. But a level of trust needs to be preserved for cameras to be deployed in privacy-sensitive settings in the home, such as toilets and bedrooms. Borrowing the categorisation of privacy provided by \cite{clarke1999internet, clarke2006s}, what is crucial, however, is the need to preserve the resident's bodily privacy in various sensitive scenarios. Bodily privacy refers to the privacy regarding images of the body. More precisely, it considers the activities that are carried out, and the loss of privacy given the nature of some of these activities (e.g., nudity during showering, etc.). What is also of interest is to preserve the privacy of sensitive personal behaviour, such as a person's political activities, sexual habits, religious practices, and with the personal space required to facilitate such behaviour. To obtain and preserve this element of trust, visual privacy needs to be preserved at every stage of a system used for monitoring.


With this idea in focus, this document surveys the state of the art in visual privacy protection methods, with special attention paid to the concept of visual obfuscation. The dichotomy between identity protection and bodily privacy can also be observed in the classification scheme this paper proposes for visual privacy preservation techniques. \textit{Perceptual obfuscation} methods (explained in Section \ref{sec:perceptual_obf}) aim to preserve trust through the protection of bodily privacy. \textit{Machine obfuscation} methods (explained in Section \ref{sec:machine_obf}) are mainly aimed at the protection of identity from machine learning models.


The issue of privacy preservation goes hand in hand with the creation of trusted pipelines, which help in increasing the adoption of monitoring technology in places like AAL care homes, where the use of such technology can be highly beneficial. This review introduces a  framework with which visual privacy protection methods can be classified under, and introduces terminology that can be used to categorise methods developed to provide visual privacy. It attempts to capture the field in a broad sense, while also connecting the state-of-the-art in the field to the framework of privacy by design \citep{cavoukian2009privacy}. This is important, since privacy is a societal problem, as opposed to being a challenge that is purely technical in nature. Solutions that are deployed need to provide privacy from the ground up, while providing users with enough knowledge and options to control the flow of data which is obtained from their actions. 

The central contributions of this review are as follows: 
\begin{enumerate}
    \item With emphasis on visual obfuscation methods, this paper reviews the state of the art in visual privacy protection methods.
    
    \item It connects low-level concepts in the field of visual privacy to high-level concepts encountered when discussing privacy by design.
    
    \item This paper proposes a novel classification scheme to make sense of visual obfuscation methods. 
\end{enumerate}

The rest of the review is structured as follows. Section \ref{sec:methodology} looks at prior reviews. Section \ref{sec:privacyProtectionMethods} explores the state of the art in visual privacy protection methods.  A novel classification scheme for the methods in this category is also introduced. Here, the review expands on those methods that are classified by the scheme under the categories of \textit{intervention methods}, \textit{blind vision}, \textit{secure processing} and \textit{data hiding} \citep{padilla2015visual}. 

Section \ref{sec:vis_obfuscation} explores in greater detail the state of the art in visual obfuscation methods, another subcategory of visual privacy protection methods that is essential to this review. In this section, the review also ties privacy preservation methods to the concept of pipelines and systems that ensure visual privacy, and to the idea of broadcasting different visualisations based on access privileges. 

Section \ref{sec:pbd} explains the concept of privacy by design, a high-level concept in systems design essential to the creation of truly end-to-end private systems. In this section, the paper links together a categorisation scheme proposed for ensuring privacy by design to the scheme proposed in this review for categorising visual privacy protection methods. This is done to link both high and low-level concepts encountered when discussing visual privacy. 

Section \ref{sec:perfeval} introduces the reader to performance evaluation setups used when measuring the efficacy of privacy preservation techniques. Important technical privacy metrics which are frequently employed are explored. It also introduces the reader to datasets that are commonly used to train models that work to impart visual privacy. Meta-studies are also explored which evaluate the real-life effectiveness of performance evaluation frameworks employed for privacy preservation techniques, through the use of user acceptance studies. Finally, Section \ref{sec:conclusion} concludes the survey by introducing the reader to important future work to be conducted to advance the field.

This work is meant to serve as more than merely a survey of the state-of-the-art. It seeks to provide the connection between high-level concepts defined in the area of privacy by design to the lower level taxonomy of methods proposed in this review. This is meant to introduce the reader to the idea of end-to-end privacy preserving systems to be used in areas like care homes. This is to highlight the practical relevance of privacy preserving technologies developed, and to push the field towards a place where more of the techniques developed through research is deployed in real-world scenarios. This is especially important when considering the ageing demographics in most developed nations, a trend that is expected to continue in the future. Considering this, there is the urgent need for more privacy to be imparted to the part of the population which is going to be residing and monitored in private settings or in care homes.

\section{Prior Reviews}
\label{sec:methodology}
Prior work has attempted to systematise knowledge in the field of visual privacy preservation, notable examples being reviews by \cite{padilla2015visual}, \cite{ribaric2016identification}, and \cite{meden2021privacy}. \cite{padilla2015visual} introduces the reader to a taxonomy of visual privacy preservation techniques seen in the literature. These are grouped under five major categories based on the manner in which they impart privacy, these being \textit{Intervention Methods}, \textit{Blind Vision}, \textit{Secure Processing}, \textit{Data Hiding}, and \textit{Redaction methods}. Redaction methods are further subdivided into \textit{image filtering, encryption, k-same family of algorithms, object / people removal}, and \textit{visual abstraction}. The authors also provide a survey of privacy-aware intelligent monitoring systems as part of their review.

\begin{figure}[t]
    \centering
    \includegraphics[width=\linewidth]{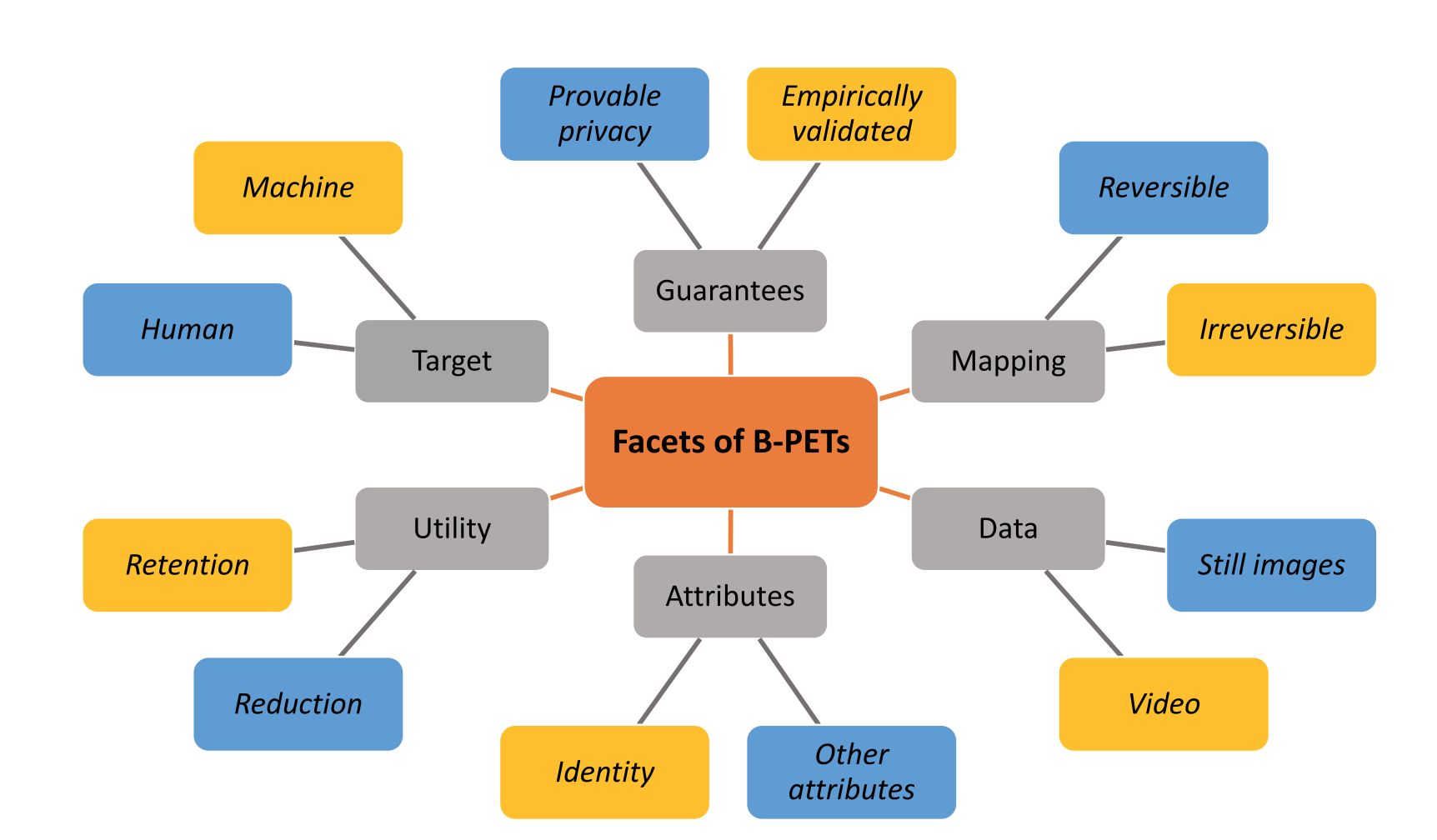}
    \caption{Classification of Biometric Privacy Enhancing Technologies (Reprinted from \cite{meden2021privacy}).}
    \label{fig:bpets}
\end{figure}

Another more recent work is by \cite{meden2021privacy}, which provides a taxonomy of methods for the area of biometric privacy enhancing technologies, paying particular attention to facial biometrics. The survey also introduces a taxonomy of biometric privacy enhancing techniques. The taxonomy of methods is grouped based on 6 criteria, namely~— \textit{the target biometric attributes}, \textit{biometric utility}, referring to the usefulness of data for automatic extraction of various attributes like health indicators and identity information, \textit{guarantees of reconstruction} from privacy-enhanced data, \textit{biometric attributes targeted}, \textit{type of mapping used}, and \textit{type of data} the method is applied to. The classification scheme introduced along with the grouping criteria can be seen in Figure~\ref{fig:bpets}.

The survey by \cite{ribaric2016identification} is a broader survey of the field of privacy preservation, touching on both visual and non-visual (e.g. audio) aspects of privacy. The survey provides an overview of de-identification approaches for non-biometric identifiers (e.g. text, hairstyle, dressing style, licence plates), physiological identifiers (e.g. face, fingerprint, iris, ear), behavioural (e.g. voice, gait, gesture) and soft-biometric (e.g. body silhouette, gender, age, race, tattoo) identifiers in a multimedia context (see figure \href{https://ars.els-cdn.com/content/image/1-s2.0-S0923596516300856-gr1.jpg}{here}). The authors then present examples of methods used to provide privacy to users based on these classifiers.

In contrast to the prior reviews in the field, this work seeks to present privacy preservation techniques that are meaningful in AAL applications. Therefore, the focus is on protecting bodily privacy, and is not concerned with whether the identity of the person is protected, as that is something commonly of a public nature. A broader exploration of the state of the art is presented, tying together concepts from the privacy by design literature to ideas coming from computer vision. 



As the focus of this review is on biometric identifiers that affect bodily privacy in the scenario of visuals from private settings or care home environments supporting AAL, some identifiers of direct importance here are behavioural identifiers (e.g., gait, gestures, actions, or activities), dressing styles, and body silhouettes. It might also be the case that wearable cameras are used to provide an AAL service. In this case, when the resident moves out of the private environment, they might encounter other persons who might not have consented to being monitored. Hence, there is a need for stricter measures of privacy to be implemented through the obfuscation of other biometric identifiers. These are faces (in still and video images); gait, and gesture; scars, marks, and tattoos; and the hair, and dressing style. These have the potential to reveal the identity of passers-by to observers of the visual feed. 


Obfuscation of some of these above-mentioned identifiers: scars, marks, and tattoos, and the hairstyle or dressing style have not been explored in the literature to the best of the authors' knowledge. Anonymisation techniques targeting other identifiers are explored in some depth in the next sections of this review, namely those concerning body silhouettes (using full-body de-identification), gait, and faces.

Papers in the field of visual obfuscation reviewed in this work are listed in Table \ref{tab:landscapeoftech}. Importance is given to research published in the field of \textit{perceptual obfuscation}, as it is especially relevant for AAL. This work also puts more emphasis on work published after 2016, as it reviews the state of the art in the field. Since the rise of deep learning, the field of computer vision has also undergone a revolutionary change. Arguably, most state-of-the-art methods proposed to impart visual privacy attempt to do so through the use of deep learning. This is also reflected in the methods surveyed as part of this review.

\section{Visual Privacy Preservation Methods}
\label{sec:privacyProtectionMethods}
\begin{figure}[t]
    \centering

\tikzset{every picture/.style={line width=0.75pt}} 

\begin{tikzpicture}[x=0.75pt,y=0.75pt,yscale=-0.45,xscale=0.5, every node/.style={scale=.65}]

\draw    (158,280) -- (199,280) -- (198,236) -- (197,167) -- (280,167) ;
\draw    (536,508) -- (564,508) -- (564,545) -- (625,545) ;
\draw    (622,457) -- (563,456) -- (564,508) -- (536,508) ;
\draw    (276,223) -- (197,222) ;
\draw    (798,496) -- (753.1,495.96) -- (754,543) -- (726,543) ;
\draw    (754,543) -- (754,580) -- (798,581) ;
\draw    (199,280) -- (198,362) -- (198,409) -- (277,410) ;
\draw    (198,341) -- (242,341) -- (281,341) ;
\draw    (199,281) -- (275,282) ;
\draw    (380,408) -- (408,409) -- (408,501) -- (443,500) ;
\draw    (381,407) -- (408,408) -- (409,237) -- (458,236) ;
\draw[dotted]   (585,286) -- (585,335) -- (633,335) ;
\draw    (562,238) -- (585,238) -- (585,151) -- (625,151) ;
\draw    (629,194) -- (586,195) ;
\draw    (628,239) -- (584,239) ;
\draw    (629,287) -- (586,286) -- (585,236.5) ;

\draw (35.39,253.58) node [anchor=north west][inner sep=0.75pt]   [align=left] {Visual Privacy\\Preservation\\Techniques };
\draw (330.78,281.17) node   [align=left] { \ \ Secure \\Processing };
\draw (284.5,391.34) node [anchor=north west][inner sep=0.75pt]   [align=left] { \ \ \ Visual\\ Obfuscation};
\draw (290.09,331.68) node [anchor=north west][inner sep=0.75pt]   [align=left] {Data Hiding};
\draw (284.96,148.1) node [anchor=north west][inner sep=0.75pt]   [align=left] {Intervention\\ \ \ Methods};
\draw (449.72,486.05) node [anchor=north west][inner sep=0.75pt]   [align=left] { \ Machine \\Obfuscation};
\draw (468.24,214.52) node [anchor=north west][inner sep=0.75pt]   [align=left] {Perceptual \\Obfuscation};
\draw (633.48,536.8) node [anchor=north west][inner sep=0.75pt]   [align=left] {Poisoning\\ Attacks};
\draw (637.41,445.14) node [anchor=north west][inner sep=0.75pt]   [align=left] {Evasion\\ Attacks};
\draw (808.31,560.03) node [anchor=north west][inner sep=0.75pt]   [align=left] {Clean Label\\ Attacks};
\draw (808.6,477.93) node [anchor=north west][inner sep=0.75pt]   [align=left] {Model Corruption\\  Attacks};
\draw (634,186) node [anchor=north west][inner sep=0.75pt]   [align=left] {Facial De-identification};
\draw (637,128) node [anchor=north west][inner sep=0.75pt]   [align=left] {Image Filtering};
\draw (636,229) node [anchor=north west][inner sep=0.75pt]   [align=left] {Total Body Abstraction};
\draw (641,317) node [anchor=north west][inner sep=0.75pt]   [align=left] {Environment \\Replacement};
\draw (640,276) node [anchor=north west][inner sep=0.75pt]   [align=left] {Gait Anonymisation};
\draw (337.95,222.5) node   [align=left] {Blind vision};

\end{tikzpicture}

\caption{A taxonomy of visual privacy preservation techniques for AAL. The topic of environmental privacy is connected with dotted lines to show that it is an under-researched but important topic.}
\label{fig:overview}
\end{figure}
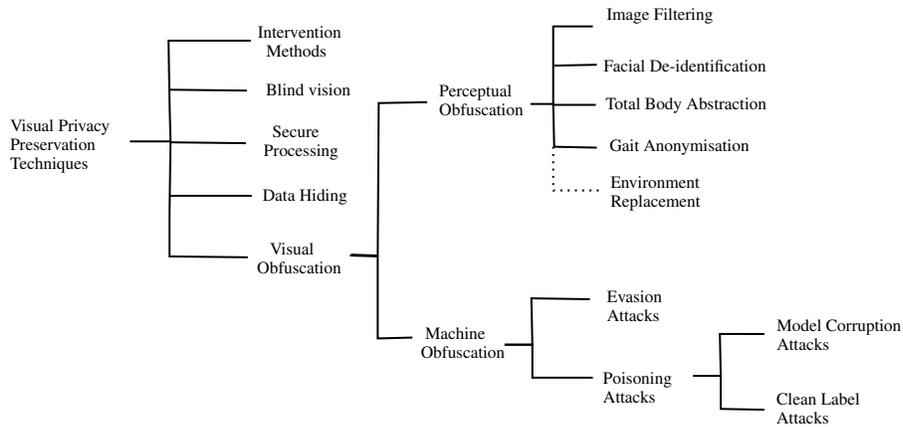

Building on the taxonomy for visual privacy preservation methods introduced by \cite{padilla2015visual}, this review categorises visual privacy preservation methods into 5 categories: \textit{intervention methods}, \textit{blind vision}, \textit{secure processing}, \textit{data hiding}, and \textit{visual obfuscation}. The taxonomy can be seen visualised in Figure~\ref{fig:overview}.

\subsection{Intervention Methods}
\label{subsec:intervention}
Intervention methods are those techniques that interfere during the data collection phase, preventing private visual data from being collected from the environment. \cite{Perez2017bystanders} classify these methods under three categories~— \textit{sensor saturation}, \textit{broadcasting commands}, and \textit{context-based approaches}.

\textbf{Sensor saturation} methods impart privacy by feeding the input device's sensor a signal that is far more in amplitude than the maximum that the device can handle. Physical interventions that prevent the capture of private images under sensor saturation schemes are also present under this category. One of the most commonly used intervention methods of this type are commercial webcam covers, also known as privacy stickers for laptops and phone cameras. These are stickers that can be stuck onto the camera, and some can be closed and opened at will. The nature of the adhesive and the construction of the blocking mechanism differs between methods \citep{jonsson2016webcam, barangan2015microphone, haddad2017detachable, miller2020electronic,mitskog2012camera, rolle2018camera}. 

The Blindspot system developed by \cite{patel2009blindspot} is an example of another intervention  of this kind. Blindspot consists of a camera lens tracking system that locates retro-reflective CCD or CMOS cameras in the vicinity, along with directing a pulsing light at the camera's lens that distorts recorded visuals. Anti-paparazzi devices have also been devised that qualify as intervention methods. \cite{harveyanti} describe anti-paparazzi devices that are cloaked as fashionable clutch bags. These detect camera flashes with the use of light sensors along with IR sensors to detect autofocus lights. The intervention device then uses an array of LEDs to produce pulses of light bright enough to overexpose photos taken by the photographers.

\cite{zhu2017automating} created the concept of LiShield, which protects a physical scene against photographing. This is achieved through the use of smart LEDs, which emit specially constructed waveforms to illuminate a scene. The LEDs emit intensity modulated waveforms that are imperceptible to the human eye, but their waveforms are constructed in such a way as to interfere with the image sensors of mobile camera devices. Mobile phones have also started to be shipped with inbuilt mechanisms for sensor saturation-based intervention. Examples include the PinePhone \citep{Pinephone} which comes with physical `kill switches' for configuring its hardware. These can be individually configured to disable both its front and rear cameras, among other peripherals \citep{PinePeripherals}. 

\textbf{Broadcasting commands} are another category of intervention methods, where devices broadcast commands using various communication protocols to disable input devices present around the subject. One example is Hewlett-Packard's concept of a paparazzi-proof camera. This includes cameras with inbuilt facial recognition, which upon receiving a remote command, selectively blurs sensitive parts of images that includes faces \citep{pilu2007detector}. Broadcasting commands are considered less effective than their physical counterparts. This is because user consent\footnote{In this case the photographer's as they are the users of the camera} is required for these methods to work. Broadcasting commands are also arguably less popular as intervention methods than sensor saturation methods.
 
\textbf{Context-based approaches} are used by devices that use various methods of context recognition to understand the scene of data collection. Once recognised, the context is used to dictate whether data is to be collected or not by triggering software actions at the sensor level. One example of this is the \textit{Virtual Walls} framework described by \cite{kapadia2007virtual}, where devices use contextual information such as GPS data to trigger software action like the disabling of sensors in the device. This allows users to control their digital footprint. To the best of the authors' knowledge, this has not been implemented in commercial devices. Context-based approaches are also arguably less popular than intervention methods. 

\subsection{Blind Vision}
\label{subsec:blindvision}
Blind vision \citep{avidan2006blind} refers to the methods by which the processing of images and videos is done in an anonymous way. The authors describe the example of two operators, Bob and Alice, to illustrate an example of blind vision. Bob possesses a face detection algorithm that he does not want compromised. Alice, on the other hand, possesses private data on which a face detection algorithm is required to be run.  Blind vision methods allow commonly used computer vision tasks to be executed without compromising on privacy of neither the algorithm used for computing, nor the data itself. In this case, both Alice and Bob can operate without compromising on the privacy requirements of either side. Blind vision works through the use of secure multi-party computation (SMC) techniques, a subfield of cryptography that allow computations to be performed privately. This allows algorithms to be executed privately, but at the same time leads to the slowdown of computation due to the overhead involved.

Some notable techniques belonging to the category of blind vision include schemes by \cite{avidan2009oblivious}, \cite{erkin2009privacy}, and \cite{sadeghi2009efficient}. 
To the best of the authors' knowledge, blind vision is not a field that is actively researched.

\subsection{Secure Processing}
\label{subsec:secureproc}
Those privacy preservation methods that are not based on SMC, but which still can process visual information in a privacy respectful way, are classified in this review under secure processing. These refer to algorithms and queries where privacy is required in a unidirectional sense: the databases on which the queries are performed are usually public, but the query and its results are to be kept private. The image matching algorithm for private content-based image retrieval (PCBIR) by \cite{shashank2008private} is one example. Algorithms that reject visual information that is not necessary for processing are also considered by the authors to be under the framework of secure processing. As an example, consider the concept of using depth or thermal cameras as the sensor device in conducting privacy preserving machine learning. These devices allow the observer to glean some information from the visual feed (e.g., number of people in the room, the activity being performed etc) while hiding the most commonly utilised privacy-sensitive information (facial identity, location information, etc)~\citep{Heitzinger2020}. 

There are also secret sharing schemes that can be classified under secure processing, wherein inference is not done on the original data, but on privacy preserving derived data obtained from the original. One example is the scheme proposed by \cite{upmanyu2009efficient}, in which images are split into multiple privacy preserving parts, which can then be distributed across nodes. Algorithms can then be applied on these image parts privately. Homomorphic encryption schemes also figure into the space of secure processing. These allow data to be encrypted in such a way that algorithms can still be run with utility on the resulting encrypted data, thereby protecting privacy. Homomorphic encryption has been successfully applied in computer vision applications as well \citep{yonetani2017privacy, bian2020ensei}. 

\subsection{Data Hiding}
\label{subsec:datahiding}
Data hiding methods refer to privacy preservation methods that, in addition to modifying privacy-sensitive regions in images, aim to embed the original information inside the modified image so that the original can be retrieved if its need arises. \cite{petitcolas1999information} provide a useful classification of data hiding methods. Under the process, embedded data (secret message) is hidden within another message (cover message) which in this case is a video frame. Thus, a marked message is obtained as a result of this hiding process. Data hiding techniques include steganography, digital watermarking, and fingerprinting. Steganography uses a key to allow the recovery of the secret message. Digital watermarking encodes the information about the ownership of an object by a visible pattern, such as a logo. Fingerprinting, conversely, hides serial numbers that uniquely identify an object inside an image, such that the owner of the copyright can detect violations of licence agreements. In the context of visual privacy protection, watermarking can be used to hide the sensitive attributes in an original video inside an obfuscated version. As an example, for facial privacy preservation, \cite{yu2007privacy} hide real faces inside frames of a video where the real face has been replaced by a generated one. Quantisation index modulation \citep{chen2001quantization} is used for the process of data hiding, and the original information can be retrieved using a secret key. This method, however, has limitation such as the artificial nature of the generated faces, and a lack of control for the generated expressions.

Depending on whether the method is fully reversible or not, data hiding techniques allow recovery of the original video to various extents. Fully reversible data hiding methods allow the original to be restored without information loss \citep{ni2006reversible}. With non-reversible methods, the original image cannot be fully restored, but this usually means an increase in hiding capacity \citep{yabuta2005new, zhang2005hiding}. 

PECAM \citep{wu2021pecam} is a method that uses elements of data hiding for creating reversible privacy-preserving transformations of images. This is, however, a method which can be used in two different modalities where the system can either be producing reversible image transformations or be irreversible. For this reason, in this review, PECAM has been categorised as a visual obfuscation method and is explained in more detail in Section \ref{sec:vis_obfuscation}.

\section{Visual Obfuscation}
\label{sec:vis_obfuscation}
This work classifies methods that seek to hide sensitive visual information directly from adversaries under visual obfuscation methods. They are divided into two major categories, \textit{perceptual obfuscation} and \textit{machine obfuscation}, based on their intention and the type of adversary from whom the private data in an image is to be obfuscated. The landscape of visual obfuscation methods analysed in this review can be seen in Table \ref{tab:landscapeoftech}.


The following sections deal with the state of the art in each of the major subcategories of perceptual obfuscation methods.

\begin{sidewaystable}[hbtp]
\begin{center}
\resizebox{\linewidth}{!}{\small 
\begin{tabular}{cccc}
\toprule 
 Category & Sub-category & Approach & Reference(s) \\
\hline 
 \multirow{23}{*}{\makecell[c]{Perceptual Obfuscation}} & \multirow{7}{*}{\makecell[c]{Image Filtering}} & Morphing & \cite{korshunov2013morph}\\ \cline{3-4} 
  &   & Warping & \cite{korshunov2013warp} \\ \cline{3-4} 
  &   & Cartooning - mean shift / adaptive filter & \cite{erdelyi2013serious}, \cite{erdelyi2014adaptive} \\ \cline{3-4} 
  &   & Cartooning - using convolutional neural networks & \cite{hasan2017cartooning} \\\cline{3-4} 
  &   & False Coloring & \cite{cciftcci2015using} \\\cline{3-4} 
  &   & PECAM & \cite{wu2021pecam} \\\cline{3-4} 
  &   & Adaptive Blurring & 
  \cite{zhang2021multi} \\ 
\cline{2-4} 
   & \multirow{3}{*}{Facial De-Identification} & Head Inpainting & \cite{sun2018natural} \\\cline{3-4} 
  &   & Live Facial de-Identification & \cite{gafni2019live} \\\cline{3-4} 
  &   & AnonymousNet & \cite{li2019anonymousnet} \\\cline{2-4} 
  & \multirow{3}{*}{\makecell[c]{Gait Anonymisation}} & Gait anonymisation using deep learning & \cite{tieu2017approach} \\\cline{3-4} 
   &   & STGAN &  \cite{tieu2019spatio} \\\cline{3-4} 
  &   & Gait anonymisation from Low quality silhouettes &  \cite{tieu2019rgb} \\
\cline{2-4} 
   & \multirow{10}{*}{Total Body Abstraction} & Generative Full Body and Face De-Identification & \citep{brkic2017know} \\\cline{3-4} 
  &   & SMPLicit & \cite{corona2021smplicit} \\\cline{3-4} 
  &   & FrankMocap & \cite{rong2020frankmocap} \\\cline{3-4} 
  &   & DensePose & \cite{neverova2018dense}\\\cline{3-4} 
  &   & Dense correspondences using depth sensors &\cite{taylor2012vitruvian}, \cite{wei2016dense} , \cite{pons2015metric}\\\cline{3-4} 
  &   & Dense correspondences using RGB images & \cite{bristow2015dense}, \cite{zhou2016learning}, \cite{gaur2017weakly} \\\cline{3-4} 
  &   & Object removal using PDE inspired algorithms & \makecell[c]{\cite{perona1990scale},\cite{bertalmio2000image}, \cite{rudin1992nonlinear}} \\\cline{3-4} 
  &   & Object Removal - Exemplar & \cite{criminisi2004region}\\\cline{3-4} 
  &   & Object Removal \ - Hybrid & \makecell[c]{\cite{bertalmio2003simultaneous},  \cite{zhang2012image}, \cite{cho2008image}} \\\cline{3-4} 
  &   & Object Removal - Deep Learning & \cite{yeh2017semantic}, \cite{yu2018generative},  \cite{kim2019deep}, \cite{chang2019free}, \cite{lee2019copy}, \cite{zhang2019internal}, \cite{oh2019onion} \\ 
\hline 
 \multirow{4}{*}{\makecell[c]{Machine \\Obfuscation}} & \multirow{2}{*}{Evasion Attacks} & Spectacles & \cite{sharif2016accessorize}  \\\cline{3-4} 
  &   & Adversarial stickers and patches  & \cite{komkov2021advhat}, \cite{brown2017adversarial}, \cite{wu2020making}, \cite{thys2019fooling} \\\cline{2-4} 
  & \multirow{2}{*}{Poisoning Attacks} & Clean label attacks & \makecell[c]{\cite{shafahi2018poison}, \cite{zhu2019transferable}, \cite{shan2020fawkes}} \\\cline{3-4}
  &   & Model Corruption & \makecell[c]{\cite{shen2019tensorclog}}\\
 \bottomrule
\end{tabular}}
\end{center}
\caption{Categorisation of visual obfuscation approaches reviewed.}
\label{tab:landscapeoftech}
\end{sidewaystable}

\subsection{Perceptual Obfuscation: Targetting Human Observers}
\label{sec:perceptual_obf}
In the case where obfuscation targets human observers, methods aim to impart visual privacy for users who wish to keep private from humans without the necessary access privileges, i.e. \textit{perceptually} (therefore, `perceptual obfuscation'). The primary objective of this category of methods is to create images in which the privacy-sensitive elements are perceptually different from the original. Although the lines are blurred between some methods, these types of techniques can broadly be split into five subcategories of methods based on the result~— Image filtering, facial de-identification, total body abstraction, gait anonymisation, and environment replacement. The latter, being an under-researched subject, is discussed in Section \ref{subsec:envpriv} in this review.

Perceptual obfuscation methods can also be either reversible in nature, where the original image can be retrieved after modification, or conversely be irreversible. A broad treatment of the classical literature in perceptual obfuscation can be seen in \cite{padilla2015visual}.

\subsubsection{Image Filters}
Image filtering is a class of perceptual obfuscation techniques that relies on the alteration/redaction of images in a way that imparts privacy to an image. Image filters can be applied globally to entire images, or to sensitive parts of images where privacy is required. The simplest forms of these filters are blurring and pixelation. 

Blurring filters slide a Gaussian kernel over an image, thereby using neighbourhood pixels to influence the values of a central pixel. The resulting image is one that has reduced resolution in the areas where the blurring filter has been applied (Figure \ref{fig:pipelinevis}f). Although widely used in applications as large as Google Maps, blurring has been shown to be ineffective for protecting identity against various deep learning-based attacks, even while appearing de-identified to human observers \citep{mcpherson2016defeating, oh2016faceless}. For pixelation, a grid of a certain size is chosen for the sensitive pixels in an image. For each box in the grid, an average colour over all the pixels within the box is calculated and assigned to each pixel within the box (Figure \ref{fig:pipelinevis}e). Pixelation has been widely used in the media, especially to obscure the identity of subjects who want to remain anonymous. These simple techniques have, however, been shown in various studies to not be robust in providing privacy \citep{newton2005preserving, korshunov2011video, mcpherson2016defeating, Menon_2020_CVPR}. Deblurring techniques have also been researched in literature \citep{Rozumnyi_2021_CVPR, Kupyn_2019_ICCV, zhang2020deblurring}. It could be posited that these techniques can also be repurposed as attacks against images obfuscated using blurring filters. Commercial tools for deblurring have also been developed \citep{knight_2021}.


Morphing and warping are filtering techniques primarily used for facial anonymisation. In morphing \citep{korshunov2013morph}, the input face is morphed into a target face (see figure \href{https://ieeexplore.ieee.org/mediastore_new/IEEE/content/media/6619590/6636596/6636641/6636641-fig-2-source-large.gif}{here}). This is done using interpolation and intensity parameters, which are used to steer the positions of the keypoints in the input face towards the target. In warping \citep{korshunov2013warp}, a set of keypoint parameters are determined using face detection techniques. These keypoints are then shifted according to a 'warping strength' parameter. The new intensity values are determined using interpolation. 



\cite{cciftcci2015using} devise false colours as a filter to impart visual privacy to images. For this method, RGB images are first converted into greyscale. Several colour palettes are devised for this step, and depending on the colour palette chosen, the pixel intensity is mapped into a pre-defined set of RGB pixel values. This false colour filter scheme also allows for reversible transformations from which the original image can be retrieved, through the storage of a difference image and a sign image. The authors also devise a secure pipeline for this purpose. Being a fairly generic method, the false colour scheme can be applied to nearly any RGB image, regardless of whether any pre-processing has been done on the image. This is also a lightweight scheme that can be used on entire images instead of on pre-specified areas of interest. The main downside of this approach, however, is that it is reversible if an attacker figures out the relationship between the false colour pixels and the real object's colours. A possible attack strategy that can be theorised is through the use of a neural network which is trained to learn these relationships. This introduces a serious threat to the security of the pipeline using the method, as there is no guarantee of protection of privacy (see figure \href{https://ieeexplore.ieee.org/mediastore_new/IEEE/content/media/6046/8207707/7982735/ciftc9-2728479-large.gif}{here}).

Adaptive blurring \citep{zhang2021multi} is an algorithm that relies on semantic segmentation masks to guide the process of blurring on videos. The model relies on two steps. The authors use DeepLab \citep{chen2017deeplab} to create segmentation masks for the privacy-sensitive parts of the video. A scale dependent Gaussian blur is devised for blurring those parts delineated by the mask. A custom strategy based on symmetry is used to guide the application of the Gaussian blur on the edges of the objects in question. After a bounding box is estimated for the object in question, the filter radius and standard deviation for the Gaussian blur kernel are set relative to the lengths of the two sides of the bounding box. 



One major downside of this approach is that the blurring parameters are determined simply by the estimated bounding box size. It does not consider factors like camera distortion and depth uncertainty. This leads to the algorithm miscalculating the amount of blurring necessary for some objects, leading to under-blurring or over-blurring of parts of images. Another downside of the approach is that commercial tools have devised attacks specifically to deblur obfuscated images \citep{knight_2021}. This significantly reduces the security of the pipeline when the method is used, as it provides no guarantees for the protection of private information.


Cartooning has been proposed multiple times in literature as a method for filtering images for privacy reasons. \cite{erdelyi2013serious}, for example, introduce a Meanshift-based method for cartooning. With this, they reduce the total number of colours and simplify the texture based on a neighbourhood pixel's property, and use edge recovery to preserve the sharpness of edges in the image. They also blur faces as part of the algorithm, and recolour parts of the image by shifting the hue as part of the final algorithm. \cite{erdelyi2014adaptive} also improve the previous work with the introduction of an adaptive filter, allowing users to determine the level of obfuscation. \cite{hasan2017cartooning} introduce a deep learning scheme for cartooning videos by which privacy-sensitive objects in videos are replaced by abstract cartoon clip art. For this, a region convolutional neural network (R-CNN) \citep{girshick2014rich} is used to get bounding boxes for the privacy-sensitive personal objects in the video. After selecting the right clip art and correcting for pose (the algorithm utilises the histogram of oriented gradients method of \cite{dalal2005histograms}), the clip art is inserted into the frame creating privacy-preserving cartooning effects (see figure \href{https://ieeexplore.ieee.org/mediastore_new/IEEE/content/media/8014302/8014734/8014909/8014909-fig-1-source-large.gif}{here}). 


Encryption methods for images exist which can also be thought of as image filtering, the results of which can be reversed by using a key. Naive encryption schemes treat the feed as textual data, thereby encrypting the entire stream. This leads to the algorithms not being effective in real-time scenarios. To solve this problem, various selective encryption schemes have been put forward. These work by only operating on a specific part of the image in question, thereby decreasing the total computation cost. Much of the classical literature in encryption is summarised in \cite{padilla2015visual}.

Devised by \cite{wu2021pecam}, PECAM is a system that allows for reversible filtering transformations through the use of data hiding. The PECAM system is built for streaming, and allows for the creation of filtered images that can then be reconstructed if such a need arises. In this scheme, depending on whether the model is aiming to reconstruct the images after transformation, different directions in the pipeline are followed. A generator (referred to as a transformer in the paper) neural network and discriminator (termed reconstructors) network are trained using the cycle-consistent GAN approach. The transformer is used to generate filtered images, and the reconstructor is used to regenerate the originals if need be.

In the pipeline that requires reconstruction, a secret key is generated that is used by the transformer and the reconstructor to guide the transformations. This is embedded into the image using data hiding (steganography) as an alpha channel. This RGBA image is then fed to the generator network, which after compression produces a filtered image that preserves privacy. This filtered image can then be broadcast to viewers. This image can then be fed to the reconstructor to create a reconstruction of the original image. In the cases where reconstruction is not necessary, a lightweight network is used as the generator, which is created through model distillation of the original network. After compression, this student network outputs the filtered image that is broadcasted to viewers. 
One disadvantage of the PECAM network is that the network could cause privacy leakage, as it might not work well when the privacy-sensitive objects are close to the camera. 

\subsubsection{Facial De-identification} 
Facial de-identification methods generate artificial faces through various means for protecting facial features from being identified. For this, it is necessary to blend the artificial faces into the original image. Traditional methods have relied on the use of the $k$-same family of algorithms for the task \citep{newton2005preserving, 1640608, 10.1007/11767831_15}.

\begin{figure}[t]
    \centering
    \includegraphics[width=0.9\linewidth]{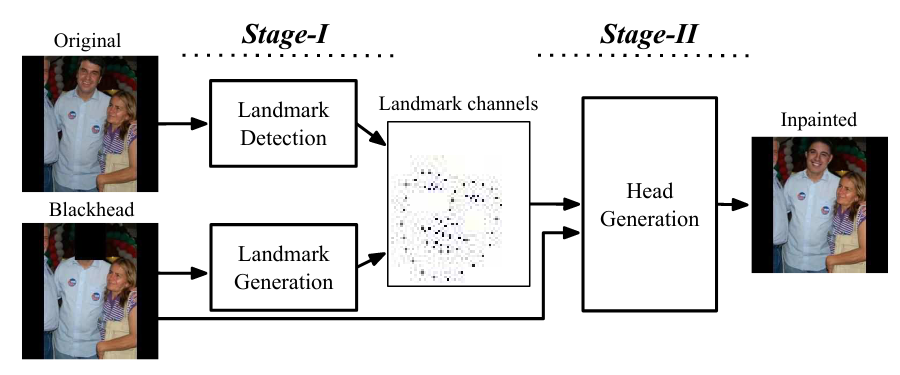}
    \caption{Two-stage facial de-identification framework used by \cite{sun2018natural}. The first stage outputs a facial landmark heatmap which is either generated or detected depending on input. This is then fed to a head generation network in the second stage along with the blackhead input image, and a generated head is inpainted into the image (Reprinted from \cite{sun2018naturalarxiv}).}
    \label{fig:sun_facial}
\end{figure}


State-of-the-art methods for facial de-identification have mostly relied on the generative power of GANs. One notable attempt is by \cite{sun2018natural}, which uses keypoint generation to condition an adversarial autoencoder (deep convolutional GANs). This scheme has two stages. The first stage uses as input either a feature-redacted blacked out or blurred image, or the original image itself. If a blacked out or blurred image is provided as input, a landmark generator is adversarially trained to accept this and generate estimates of facial landmarks in the form of a landmark heatmap. If the original image is provided, a landmark detector is used to extract the landmark heatmap required for the subsequent stage. Stage 2 accepts as input the landmark heatmap from the previous stage concatenated with the blacked out original. This is then fed to another adversarial DCGAN autoencoder that generates images in which realistic looking generated faces have been inpainted. The de-identification framework used can be seen in Figure \ref{fig:sun_facial}. 

\begin{figure}[t]
    \centering
    \includegraphics[width=0.9\linewidth]{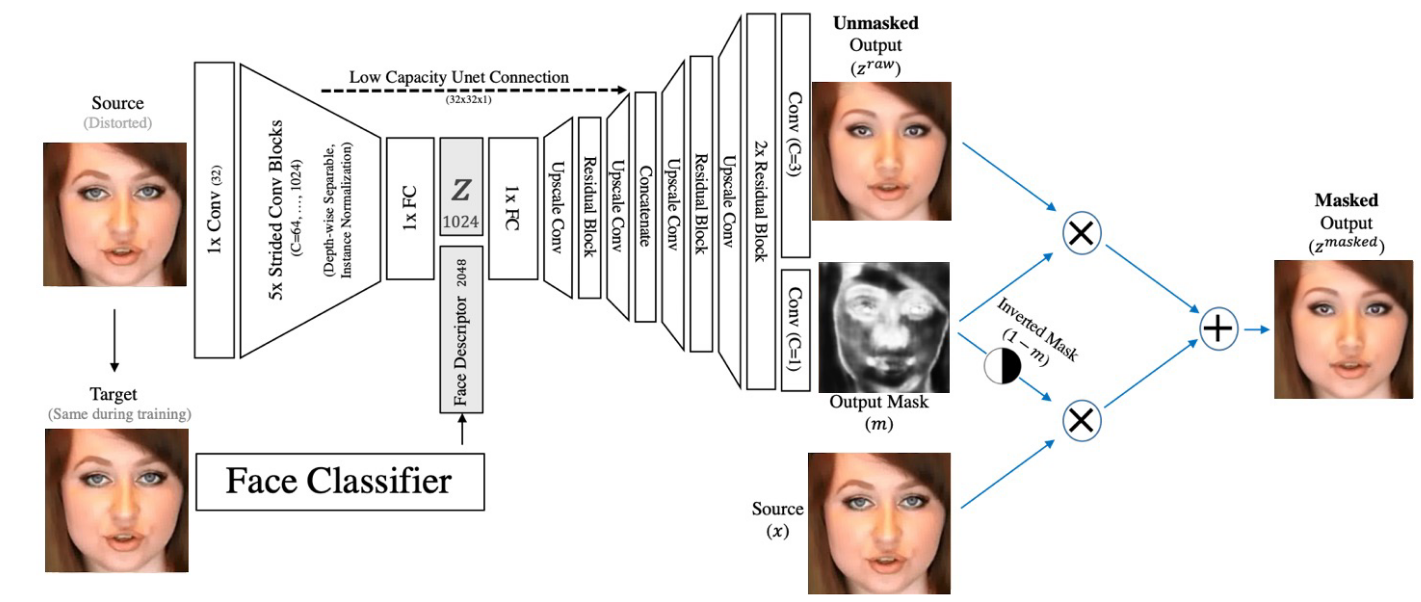}
    \caption{Framework used by \cite{gafni2019live}. The setup outputs a de-identified facial image with a similar pose, illumination, and expression to the original (Reprinted from \cite{gafni2019livearxiv}).}
    \label{fig:gafni_live_facial}
\end{figure}

\cite{gafni2019live} create a live facial de-identification method for use in videos. The system works by trying to distance the facial descriptors of a person from a target image of the person already provided to the system. This target image can be any image of the person and need not be from the same video stream. For this method, facial bounding boxes are first extracted from the video frame, and facial keypoints are extracted from this setup. A transformation matrix is obtained from this using a similarity transformation to an averaged face. This is used to then transform the input face and passed through the network to obtain an output facial image and a mask. An adversarial autoencoder network is devised for this task, and it provides a recreation of the input image, along with providing an output mask that can be used to guide the network's output. The inverse of the similarity transformation is used to transform back this output face image and a mask. A linear per-pixel mixing of the input image and the output image is done, weighted by the transformed mask. This is then merged into the original frame using the convex hull of facial keypoints, to get the final generated facial output. The framework used for de-identification can be seen in Figure \ref{fig:gafni_live_facial}.


The approach by \cite{li2019anonymousnet} is interesting for the way it straddles the worlds of both perceptual obfuscation and machine obfuscation (explored in Section \ref{sec:machine_obf}). This method, named AnonymousNet, creates perceptually altered images based on knowledge of both the facial attributes of persons observed and the distribution of those attributes in the real world (approximated by the dataset in the image). The method aligns and crops faces using a neural net referred to by the authors as a \textit{deep alignment network}, after which it does facial feature extraction using GoogleNet \citep{szegedy2015going} and random forest models \citep{breiman2001random}. This is then used as input to a custom privacy preserving attribute selection algorithm, which obfuscates the features of the face and lets the outputs resemble the features of the real world in terms of their distribution. A de-identified face is then generated by a starGAN \citep{choi2018stargan} model, conditioned by the features selected by the algorithm in the previous step. Finally, to obfuscate the outputs from machines, adversarial perturbation is done on the output image, using a universal perturbation vector defined by the DeepFool algorithm \citep{moosavi2016deepfool}. 



\subsubsection{Total Body Abstraction}
Total body abstraction methods aim to impart privacy by replacing the entire body of the subject in a visual with another generated one. Most methods under this category arguably use semantic segmentation methods to segment out humans from frames, and then subsequently replace these with abstractions such as avatars. Other visual abstractions include silhouettes, where a binary mask of the person is obtained (and sometimes modified for various purposes); invisibility, where inpainting techniques are used to replace the person with the environment/background \citep{climent2021protection}; and background subtraction, where a background image is generated and subtracted from the current frame to obtain a mask of the foreground object (here a person) of interest \citep{mondejar2019end, Rezaei2020GLBMGL}. 

One particularly interesting total body abstraction method relied on the use of generative adversarial models to generate full-body replacements. The approach by \cite{brkic2017know} use conditional GANs (DCGANs) to synthesise entire bodies of subjects, while the faces are generated using deep convolutional GAN models. The conditional GAN was trained on pairs of segmentation masks and images, and is trained to operate on segmentations with different levels of detail, from simple silhouette blobs to full-body segmentations with detailed tags for individual garments (see figure \href{https://ieeexplore.ieee.org/mediastore_new/IEEE/content/media/8014302/8014734/8014907/8014907-fig-8-source-large.gif}{here}).



State-of-the-art motion capture methods relying on the fitting of 3D avatars to humans in frames can also serve to impart visual privacy. These mostly build on the Skinned Multi-Person Linear (SMPL) model \citep{loper2015smpl}. SMPL is created to be fast and to operate with standard rendering engines, producing realistic looking avatars that do not produce the unnatural joint deformation effects commonly seen in other avatar fitting schemes. Blend shapes are represented in the scheme as a vector of concatenated vertex offsets. An artist created mesh of 6890 vertices and 23 joints is obtained. The mesh used for the rendering uses the same topology for men and women. The model also comes with other options such as a spatially variant resolution and a skeletal rig. SMPL is, however, a function solely of joint angles and face parameters. It does not consider other bodily actions such as breathing, facial motions or actions, muscle tension, or changes independent of skeletal joint angles and overall shape. SMPL also does not generalise well to account for all the variations found in people's body shapes, and produces unnatural deformations of blend shapes.

A recent example of a method devised using SMPL is Frankmocap \citep{rong2020frankmocap}, capable of both hand and body capture and replacement in real time. Since hands are harder to motion capture than most parts of the body as they are small, the authors also built a custom 3D monocular hand capture method that uses the hand part of the SMPL model to achieve this task. One drawback of this scheme is that garments are not modelled for the avatar. 

Most advancements in avatar fitting have focussed solely on returning the SMPL parameters which stand in for the 3D body meshes, ignoring the garments worn. Some advances over the standard SMPL model have focused on modelling garments worn by the person. One such recent model is the SMPLicit \citep{corona2021smplicit}. This approach specifically models garment topologies on top of the SMPL model. Garments are predicted through the use of a semantically interpretable latent vector. The objective is to then be able to influence the looks of garments by manipulating this interpretable vector. SMPL-X \citep{Pavlakos_2019_CVPR} is another extension of the SMPL model, which generates avatars with fully articulated hands and facial expressions. The Sparse Trained Articulated Human Body Regressor (STAR, \cite{STAR:2020}) improves the SMPL by producing more realistic deformations, and with only 20\% of the model parameters required for the SMPL. The model also generalises better to account for the variations in the body shapes of the human population.

The creation of a dense correspondence between images and surface-based representations is another active area of research. Some works have utilised depth images \citep{taylor2012vitruvian, wei2016dense, pons2015metric}, and others have employed RGB images to correspond to objects \citep{bristow2015dense, zhou2016learning, gaur2017weakly}. 

One noteworthy example using RGB images is the DensePose \citep{neverova2018dense} framework. The authors set about annotating persons appearing in the COCO dataset \citep{lin2014microsoft} through the use of human annotators utilising a novel annotation pipeline, thereby creating a 'DensePose-COCO' dataset. They then set about training deep neural networks to learn the associations between RGB image pixels and the surface points of human bodies. The authors use a Mask-RCNN segmentation model \citep{he2017mask} and couple it with a Dense regression system (DenseReg) \citep{alp2017densereg} for the task. DensePose has also been successfully employed in protecting visual privacy in AAL settings. \cite{climent2021protection} create various privacy preserving visualisations using a union of masks obtained from DensePose and a Mask-RCNN model, along with the original RGB image used as input for the models (See Figures \ref{fig:pipeline} and \ref{fig:pipelinevis}).

\textbf{Object/People Removal}~— 
There exists several algorithms which modify frames in a way that privacy-sensitive objects and persons in the frames are removed. These also fall under total body substitution methods. The object or person of interest inside frames, once removed, leaves a gap in the frame. This is then substituted with a generated background to create a coherent image. Inpainting methods are utilised to perform this substitution. Although methods vary, information from surrounding areas is mostly used in filling in the missing areas in the case of image inpainting methods. Considering video inpainting, information from previous frames can be utilised to perform inpainting in a subsequent frame, but temporal consistency between frames has to be ensured. This is also commonly referred to in literature as background modelling or background subtraction.

There are various techniques that have been created for image inpainting. \cite{paunwala2018image} classifies these into \textit{partial differential equation-based methods}, \textit{exemplar-based methods}, and \textit{hybrid methods}. The review introduces a category of \textit{deep learning based inpainting schemes}, which have been increasingly used since the creation of generative adversarial networks.

\textit{PDE-inspired algorithms}~— Algorithms in this category utilise geometric information to do inpainting of the gaps, by looking at the image inpainting process as one of heat diffusion. Several types of PDE-inspired algorithms exist, notably anisotropic diffusion \citep{perona1990scale}, diffusion-based image inpainting \citep{bertalmio2000image}, and total variational inpainting \citep{rudin1992nonlinear}. 

\textit{Exemplar-based methods}~— Initially created by \citep{criminisi2004region}, algorithms under this category gather information from nearby regions of the same image or from a database of images for inpainting. Texture synthesis methods can be regarded as a subsection of exemplar-based methods. In this scheme, synthetic textures derived from one portion of the image are used to fill the missing regions in another portion of the image. Texture synthesis algorithms are, however, slower than other patch-based exemplar-based methods since they do inpainting on a pixel-by-pixel basis.



\textit{Hybrid Approaches}~— Hybrid approaches combine the advantages of both PDE-based methods and exemplar-based methods to create better inpainting results. Examples include the approach by \cite{bertalmio2003simultaneous}, and the wavelet decomposition-based methods by \cite{zhang2012image} and \cite{cho2008image}

\textit{`Deep Learning'-based methods}~— 
Although their use in the scenario of object removal is scarce, deep learning models have increasingly been used for image inpainting tasks.
These typically make use of generative adversarial networks, to create realistic looking inpainted results \citep{yeh2017semantic, yu2018generative}. 
Similar approaches which have also utilised deep learning to do video inpainting include \cite{kim2019deep, chang2019free, lee2019copy, zhang2019internal, oh2019onion}. 


\subsubsection{Gait Anonymisation}
Research has pointed to the notion that gait is an important biomarker that can be used to identify individuals, as it is individually unique \citep{wang2003silhouette, bashir2010gait, liu2006improved, zhang2004human, makihara2006gait, bobick2001gait}. A deeper treatment of the subject of gait recognition can be seen in the work by \cite{wan2018survey}. Gait anonymisation is, however, a relatively newer area of research. Typically, video surveillance anonymisation tools use filters such as pixelation and blurring, and then assume the gait to be anonymised in the process \citep{agrawal2011person}. These approaches typically result in the video looking artificial and thus increase its chances of being detected as a fake. The apparent non-robust nature of classical obfuscation approaches like blurring and pixelation through targeted attacks also leave the obfuscated footage in a vulnerable position. 


Work done by \cite{tieu2017approach} proposes the use of deep neural networks to generate an anonymising gait. This gait is generated by using the original gait from the frames of the visual feed along with a specially created `noise gait' as inputs to a convolutional neural network. This network then outputs an anonymising contour vector, which after processing produces the anonymised gait. This is then placed back into the original scene. 

With the rise of generative adversarial models capable of state-of-the-art generative capabilities, newer literature has focussed on leveraging their power to produce anonymised gaits. \cite{tieu2019spatio} create spatio-temporal generative models that can obfuscate gaits present in videos, creating natural-looking sequences. This architecture makes use of one generator and two discriminators. The generator accepts the original gait and random noise to generate anonymised gaits. The first discriminator is a spatial discriminator which accepts a contour vector extracted from frames of the gait, and tries to distinguish the shape of real gaits from generated gaits at each frame. The results improve the naturalness of the shape of the generated gait. The second discriminator is a temporal discriminator, which distinguishes between the temporal continuity of the real gait and a generated gait. This determines whether the generated gait moves smoothly. A contour sequence is fed through a long short-term memory network \citep{hochreiter1997long}, the outputs of nodes of which are concatenated to form one input vector for the network. A binary anonymised gait is obtained through the generation process, which is then colourised to merge into the original background. 

This process is known to work only on high-quality silhouette inputs, and fails notably with low-quality silhouettes. \cite{tieu2019rgb} expand on this work by creating a colourisation network, in addition to a different STGAN-based generator-discriminator architecture defined in \cite{tieu2019spatio}. Through this approach, the authors were able to provide gait anonymisation for low-quality silhouettes as well (see figure \href{https://ieeexplore.ieee.org/mediastore_new/IEEE/content/media/8989870/9023008/9023188/418-fig-1-source-large.gif} {here}).


\subsection{Machine Obfuscation: Targetting Algorithms}
\label{sec:machine_obf}
This review classifies those algorithms that seek to protect the privacy of users from machine learning algorithms, under the category of machine obfuscation techniques\footnote{named so to be in contrast to the concept of machine vision}. This is currently a highly active field of research, and a large amount of work has gone towards creating machine obfuscation models. These type of algorithms, commonly categorised as attacks\footnote{because they seek to attack the validity of deep learning models used for automated analysis}, employ generative models, specifically generative adversarial networks (GANs). 

Machine obfuscation attacks can be split into two different types~— \textit{Poisoning attacks} and \textit{Evasion attacks} \citep{shan2020fawkes}. The objective of machine obfuscation attacks are to create changes in images that cause misclassification in machine recognition models. These changes are also most often imperceptible, to evade humans from detecting their presence, and also to be perceptually pleasing as to be useful for sharing on popular photo sharing applications. 

\subsubsection{Poisoning Attacks}
A subcategory of machine obfuscation attacks, termed as poisoning attacks, aims to disrupt the training of machine learning models through the introduction of specific `poisoned' images. After the introduction of such images to the set, models trained on these images behave in unexpected ways. Poisoning attacks can be further split into `clean label' attacks and `model corruption' attacks.

\textbf{Clean Label Attacks}~—
In clean label attacks, adversarial noise is created so that models trained on the data learn to misclassify a specific image, or a set of images containing the person \citep{shafahi2018poison, zhu2019transferable}. This is done by creating the adversarial noise in a very specific way as to alter the feature space that is used by machine learning models for recognition.  In the test phase, after encountering an unaltered image, a model classifies the image incorrectly due to it seeing a different feature vector than what was seen during training. 

Most clean label attacks work on the possible misclassification of a single preselected image that is introduced, although exceptions do exist. \cite{shan2020fawkes} developed \textit{Fawkes}, which is one such approach through which users can produce `cloaked' images of themselves through the addition of imperceptible adversarial noise. These then cause machine learning models trained on the cloaked images to misclassify normal images of the user. 



\textbf{Model Corruption Attacks}~—
A model corruption attack aims to distort the feature space of images in such a way that upon using the altered images, it reduces the overall accuracy of the trained model \citep{shen2019tensorclog}. The objective of model corruption attacks are to prevent unauthorised data collection and model training. One disadvantage of these types of attacks is that they are more easily detectable because the presence of such an attack would be readily reflected in the drop in overall model accuracy seen.


\subsubsection{Evasion Attacks}
Evasion attacks create images that are difficult for image recognition systems to identify. These commonly rely on the creation of adversarial examples through the use of real-life artefacts, which upon being shown to cameras during capture increases the chances of the subject being misidentified. Prominent examples of this sort include wearables like a specially crafted pair of spectacles \citep{sharif2016accessorize}, adversarial stickers \citep{komkov2021advhat} (see figure \href{https://ieeexplore.ieee.org/mediastore_new/IEEE/content/media/9411940/9411911/9412236/9412236-fig-1-source-large.gif}{here}), or adversarial patches \citep{brown2017adversarial, wu2020making, thys2019fooling} that increase the chances of misidentification. The downside of these types of attacks is that these are obvious to a human observer of the footage. Techniques that use adversarial models to alter faces to avoid detection can also be classified under evasion attacks, while in this survey, these are moved to perceptual obfuscation techniques as they alter the appearance of the person in obvious ways, and are usually primarily aimed at human adversaries. The lines are blurred, however, as they can be created to fool machine recognition systems as well. 

Evasion attacks are not to be confused with intervention methods. While evasion attacks prevent machine learning algorithms from recognition through the use of hardware, these do not prevent the collection of the data itself. Intervention methods, on the other hand, use specialised hardware to interfere during the data collection stage, preventing private data from ever being sent to the subsequent stages of the pipeline.

\begin{figure}[t]
    \centering
    \includegraphics[width=0.9\linewidth]{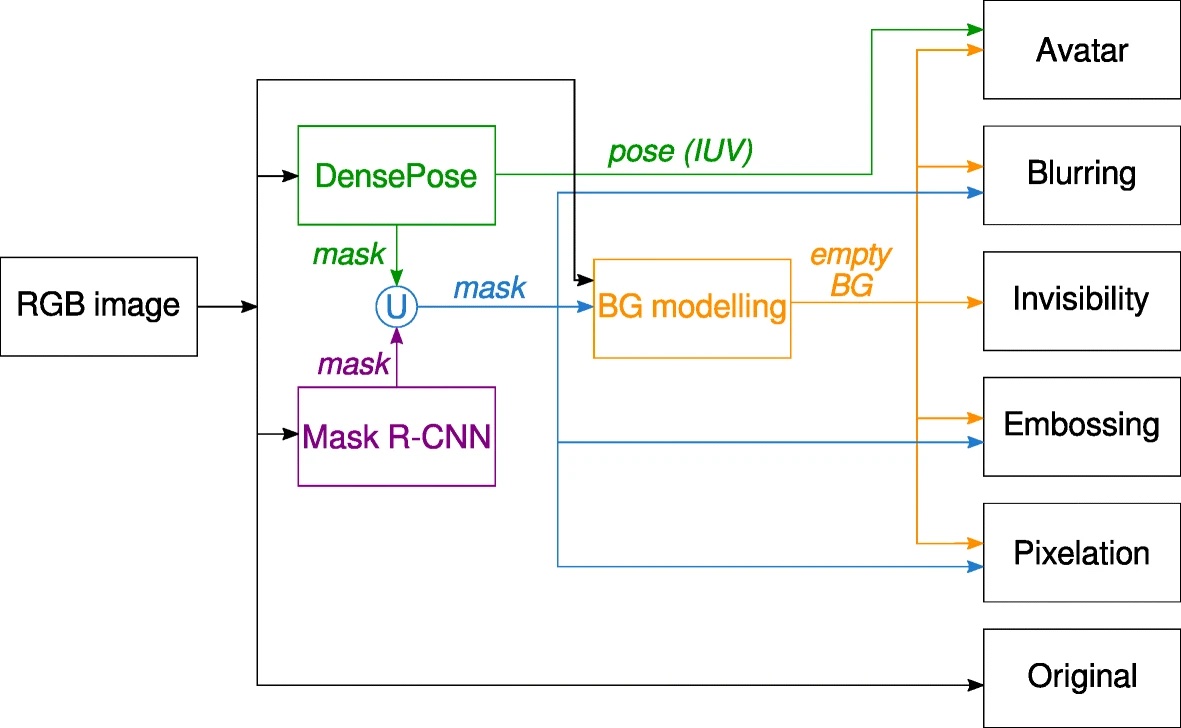}
    \caption{An illustration of a pipeline that accepts RGB images and applies various privacy preserving filters according to access privileges (reprinted from \cite{climent2021protection}).}
    \label{fig:pipeline}
\end{figure}

\subsection{Privacy Protecting Pipelines}
\begin{figure}
    \centering
    \includegraphics[width=0.8\linewidth]{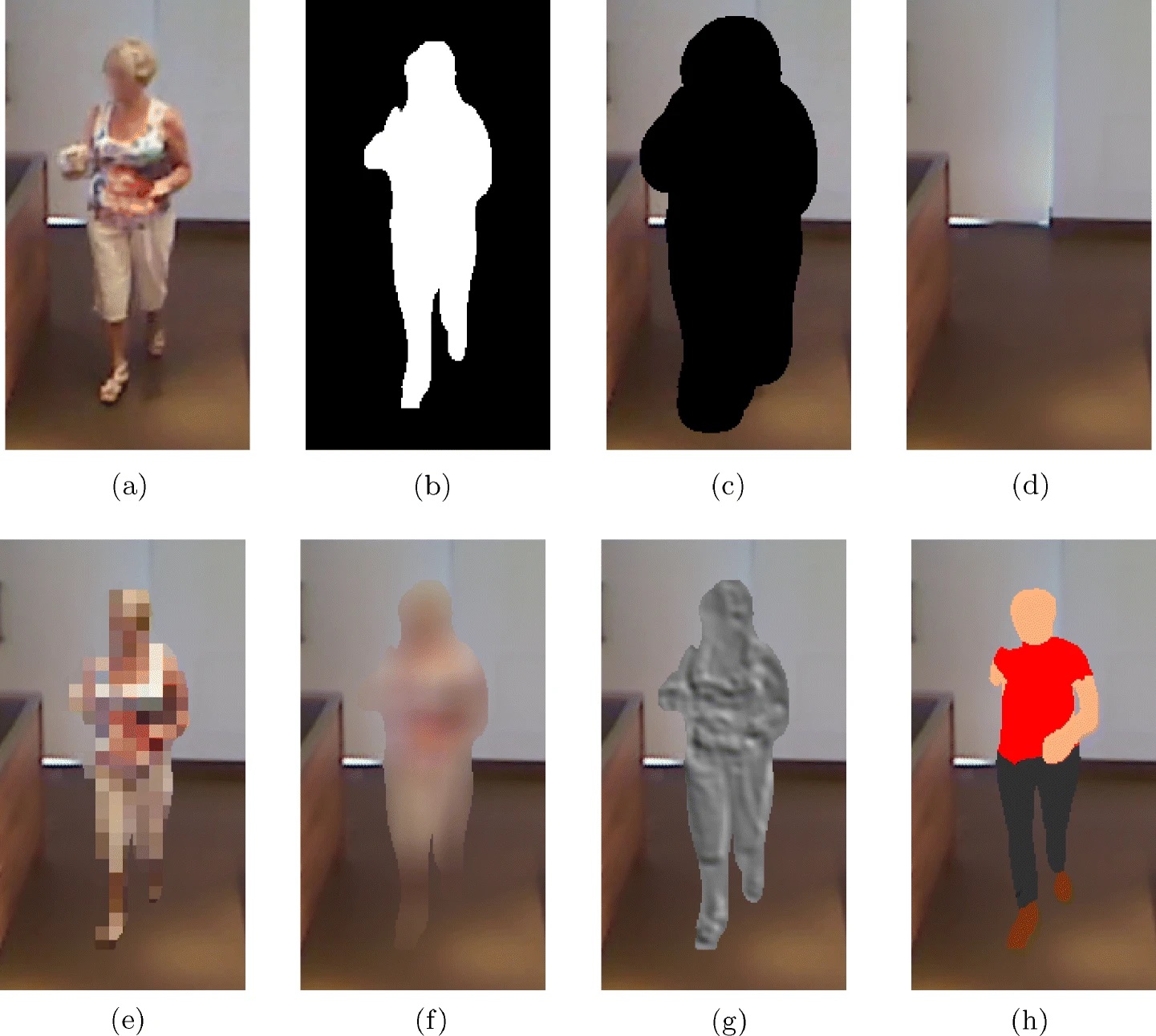}
    \caption{Example frame from the Toyota Smarthome dataset, within the workflow of the method proposed in \cite{climent2021protection}. (a) shows the original frame; (b) shows the union mask obtained for this frame; (c) shows the background image fed to the background updating scheme; (d) through (h) show results after applying the different filters (in order~— invisibility, pixelation, blurring, embossing, and avatar). (Reprinted from \cite{climent2021protection}).}
    \label{fig:pipelinevis}
\end{figure}

Research has also been conducted to create end-to-end pipelines that aim to preserve visual privacy through the combination of various techniques in visual privacy preservation. One notable example is by \cite{climent2021protection} (see Figure~\ref{fig:pipeline}). Here, the authors accept an RGB image as input, creating with it a Densepose \citep{neverova2018dense} and Mask R-CNN \citep{he2017mask} masks. Using these representations along with a background model created after using a union of the two masks as input, the authors produce five privacy preserving representations, namely the avatar, blurring, invisibility, embossing, and pixelation. These preserve privacy to differing extents, and the footage can be broadcast to users depending on access privileges. The results from the application of the pipeline on a frame from the Toyota Smarthomes dataset \citep{das2019toyota} can be seen in Figure~\ref{fig:pipelinevis}.

\section{Privacy by Design}
\label{sec:pbd}
Privacy by Design is a systems design concept defined by \cite{cavoukian2009privacy}, which advances the view that privacy cannot be ensured through compliance with regulatory frameworks, and must instead stem from an organisation's default mode of operation. The concept is accomplished through adhering to the following 7 principles:

\begin{enumerate}
    \item Proactive not Reactive; Preventative not Remedial~— Systems ought to be created that prevent privacy invasive events before they occur.
    \item Privacy as the Default Setting~— In any business practice or IT system, an individual's privacy is automatically protected even if they perform no actions. 
    \item Privacy Embedded into Design~— Privacy is embedded into the core design and architecture of IT systems, and into the surrounding business practices.
    \item Full Functionality (Positive-Sum, not Zero-Sum)~— False dichotomies, such as that of privacy vs security, is avoided. The system seeks to accommodate legitimate interests of both the user and service provider in a win-win fashion.
    \item End-to-End Security (Full Lifecycle Protection)~— The system architecture ensures that strong security measures which are essential to ensuring privacy are established, extending through the entire lifecycle of the data. 
    \item Visibility and Transparency (Keep it Open)~— Components of the system are created in a way as to be visible and transparent to users and data providers. This ensures verification of the objective that the business is operating according to its stated promises.
    \item Respect for User Privacy (Keep it User-Centric)~— The system is architected in such a way that the interests of the individual is upheld. This is done through providing strong privacy defaults, appropriate notice, and user-friendly options.
\end{enumerate}


Based on different design elements present in lifelogging technologies, \citep{mihaildis2020methodological} created a classification schema that separates privacy by design into levels. According to the schema, components in a pipeline acting at each level must be compliant with existing data protection rules for the system to adhere to the notion of privacy by design.

The most basic of these is at the \textit{sensor level}. Moving upwards in scope, they can be specified as \textit{model level}, \textit{system level}, \textit{user interface level}, and at the most abstract, privacy at the \textit{user level}. For clarity, this is connected to the taxonomy of visual privacy preservation methods presented in Section~\ref{sec:privacyProtectionMethods}. The correspondence between both taxonomies can be seen in Figure~\ref{fig:PrivByDesign}, and is further explained in subsequent subsections. 

\usetikzlibrary{patterns}
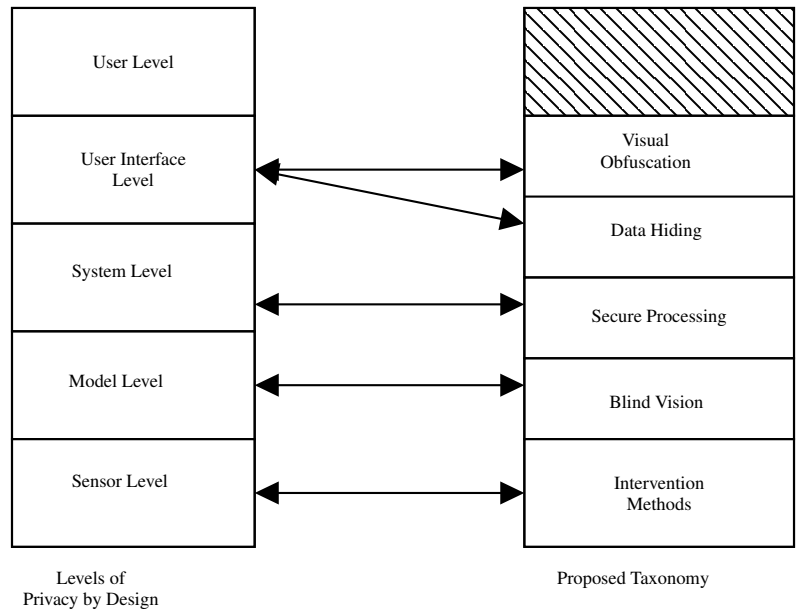
\begin{figure}[t]
\centering

\tikzset{
pattern size/.store in=\mcSize, 
pattern size = 5pt,
pattern thickness/.store in=\mcThickness, 
pattern thickness = 0.3pt,
pattern radius/.store in=\mcRadius, 
pattern radius = 1pt}
\makeatletter
\pgfutil@ifundefined{pgf@pattern@name@_1qczr7z6g}{
\pgfdeclarepatternformonly[\mcThickness,\mcSize]{_1qczr7z6g}
{\pgfqpoint{0pt}{-\mcThickness}}
{\pgfpoint{\mcSize}{\mcSize}}
{\pgfpoint{\mcSize}{\mcSize}}
{
\pgfsetcolor{\tikz@pattern@color}
\pgfsetlinewidth{\mcThickness}
\pgfpathmoveto{\pgfqpoint{0pt}{\mcSize}}
\pgfpathlineto{\pgfpoint{\mcSize+\mcThickness}{-\mcThickness}}
\pgfusepath{stroke}
}}
\makeatother
\tikzset{every picture/.style={line width=0.8pt}} 

\begin{tikzpicture}[x=0.85pt,y=0.85pt,yscale=-1.2,xscale=1.2, every node/.style={scale=1}]

\draw   (50,20) -- (140,20) -- (140,220) -- (50,220) -- cycle ;
\draw   (50,20) -- (140,20) -- (140,60) -- (50,60) -- cycle ;
\draw   (50,60) -- (140,60) -- (140,100) -- (50,100) -- cycle ;
\draw   (50,100) -- (140,100) -- (140,140) -- (50,140) -- cycle ;
\draw   (50,140) -- (140,140) -- (140,180) -- (50,180) -- cycle ;
\draw   (50,180) -- (140,180) -- (140,220) -- (50,220) -- cycle ;
\draw   (240,20) -- (340,20) -- (340,220) -- (240,220) -- cycle ;
\draw  [pattern=_1qczr7z6g,pattern size=6pt,pattern thickness=0.75pt,pattern radius=0pt, pattern color={rgb, 255:red, 0; green, 0; blue, 0}] (240,20) -- (340,20) -- (340,60) -- (240,60) -- cycle ;
\draw   (240,60) -- (340,60) -- (340,120) -- (240,120) -- cycle ;
\draw   (240,120) -- (340,120) -- (340,150) -- (240,150) -- cycle ;
\draw   (240,150) -- (340,150) -- (340,180) -- (240,180) -- cycle ;
\draw   (240,180) -- (340,180) -- (340,220) -- (240,220) -- cycle ;
\draw    (240,90) -- (340,90) ;
\draw    (143,200) -- (237,200) ;
\draw [shift={(240,200)}, rotate = 180] [fill={rgb, 255:red, 0; green, 0; blue, 0 }  ][line width=0.08]  [draw opacity=0] (8.93,-4.29) -- (0,0) -- (8.93,4.29) -- cycle    ;
\draw [shift={(140,200)}, rotate = 0] [fill={rgb, 255:red, 0; green, 0; blue, 0 }  ][line width=0.08]  [draw opacity=0] (8.93,-4.29) -- (0,0) -- (8.93,4.29) -- cycle    ;
\draw    (143,80) -- (237,80) ;
\draw [shift={(240,80)}, rotate = 180] [fill={rgb, 255:red, 0; green, 0; blue, 0 }  ][line width=0.08]  [draw opacity=0] (8.93,-4.29) -- (0,0) -- (8.93,4.29) -- cycle    ;
\draw [shift={(140,80)}, rotate = 0] [fill={rgb, 255:red, 0; green, 0; blue, 0 }  ][line width=0.08]  [draw opacity=0] (8.93,-4.29) -- (0,0) -- (8.93,4.29) -- cycle    ;
\draw    (142.94,80.59) -- (200.5,92.1) -- (237.06,99.41) ;
\draw [shift={(240,100)}, rotate = 191.31] [fill={rgb, 255:red, 0; green, 0; blue, 0 }  ][line width=0.08]  [draw opacity=0] (8.93,-4.29) -- (0,0) -- (8.93,4.29) -- cycle    ;
\draw [shift={(140,80)}, rotate = 11.31] [fill={rgb, 255:red, 0; green, 0; blue, 0 }  ][line width=0.08]  [draw opacity=0] (8.93,-4.29) -- (0,0) -- (8.93,4.29) -- cycle    ;
\draw    (143,130) -- (237,130) ;
\draw [shift={(240,130)}, rotate = 180] [fill={rgb, 255:red, 0; green, 0; blue, 0 }  ][line width=0.08]  [draw opacity=0] (8.93,-4.29) -- (0,0) -- (8.93,4.29) -- cycle    ;
\draw [shift={(140,130)}, rotate = 0] [fill={rgb, 255:red, 0; green, 0; blue, 0 }  ][line width=0.08]  [draw opacity=0] (8.93,-4.29) -- (0,0) -- (8.93,4.29) -- cycle    ;
\draw    (143,160) -- (237,160) ;
\draw [shift={(240,160)}, rotate = 180] [fill={rgb, 255:red, 0; green, 0; blue, 0 }  ][line width=0.08]  [draw opacity=0] (8.93,-4.29) -- (0,0) -- (8.93,4.29) -- cycle    ;
\draw [shift={(140,160)}, rotate = 0] [fill={rgb, 255:red, 0; green, 0; blue, 0 }  ][line width=0.08]  [draw opacity=0] (8.93,-4.29) -- (0,0) -- (8.93,4.29) -- cycle    ;

\draw (71,192) node [anchor=north west][inner sep=0.75pt]  [font=\scriptsize] [align=center] {Sensor Level};
\draw (70,155) node [anchor=north west][inner sep=0.75pt]  [font=\scriptsize] [align=center] {Model Level};
\draw (71,114) node [anchor=north west][inner sep=0.75pt]  [font=\scriptsize] [align=center] {System Level};
\draw (95,80) node  [font=\scriptsize] [align=center] {User Interface\\ Level};
\draw (95,40) node  [font=\scriptsize] [align=left] {User Level};
\draw (285,73) node  [font=\scriptsize] [align=center] {Visual\\Obfuscation};
\draw (289,103) node  [font=\scriptsize] [align=left] {Data Hiding};
\draw (290,135) node  [font=\scriptsize] [align=left] {Secure Processing};
\draw (289,166) node  [font=\scriptsize] [align=left] {Blind Vision};
\draw (290,200) node  [font=\scriptsize] [align=center] {Intervention\\Methods};
\draw (53,228) node [anchor=north west][inner sep=0.75pt]  [font=\scriptsize] [align=center] {Levels of \\ Privacy by Design};
\draw (251,228) node [anchor=north west][inner sep=0.75pt]  [font=\scriptsize] [align=left] {Proposed Taxonomy};

\end{tikzpicture}

\caption{Connection between the levels of Privacy by Design \citep{mihaildis2020methodological} and visual privacy protection methods.}
    \label{fig:PrivByDesign}
\end{figure}

\subsection{Sensor level}
Sensor level privacy preservation techniques prevent the capture of sensitive data in visual feeds using various software and hardware implements. These mechanisms can prevent the capture of sensitive content in the first place by the camera. This can also be implemented at the software level, as a filter to clear the captured images of protected content before the images are stored to disk. Intervention methods (Section \ref{subsec:intervention}) can be grouped under the umbrella of intervention methods, as these intervene during the data collection phase to protect the privacy of users and environments.

\subsection{Model Level}
To observe model level privacy, methods are created that preserve privacy for users while at the same time enabling models to infer information from data. Also termed as privacy-preserving data mining (PPDM), these techniques aim to create privacy in such a way that unintended third parties cannot make sense out of protected attributes in data, while also removing sensitive knowledge that has been mined from the data. 

Since blind vision methods (see Section \ref{subsec:blindvision}) help in processing the data securely, these schemes can be considered under model level methods, as they contribute to the model level privacy of the pipeline.
Since blind vision techniques also allow inferring from data while preserving privacy, it could also be noted as contributing to the system level privacy of a pipeline. Another example of a technique that contributes to the model level privacy of a pipeline is \textit{federated learning} \citep{konevcny2016federated}, a technique used for the private training of machine learning models.

\subsubsection{Federated Learning}
Machine learning, being data hungry, requires large amounts of training data. Distributing the training of models aims to speed up the process. Common distributed training schemes revolve around a central model, which is then fed with data that is kept across several nodes. This requires the transporting of data to a central location such as a data centre where the model is located, which gives rise to numerous privacy concerns.

Federated learning is a model training framework that alleviates this issue. Instead of bringing the data to the model, federated learning distributes copies of the central model to where the distributed data stores are located. These are most often user's mobile devices, as mobile computing is where federated learning is commonly used. Model training is done on the devices, and the gradients are then transferred to the main model, which is then updated. Thus, data does not leave the user's device, alleviating privacy concerns.

The ideal use cases for federated learning is when data is privacy-sensitive, and not representative of public datasets available for model training. For example, predictive keyboards need to adapt according to specific user preferences. Training on a public dataset like Wikipedia articles would lead to bad predictions. This also facilitates better training because users can correct predictions, which then yield better targets for model training. This is beneficial feedback, which is required for federated learning. This also allows models to evolve with the evolution of language. Poor use cases for federated learning are such as when the ideal dataset for training is already inside the data centre.

Privacy leakages can happen while the gradients are being sent back up to the central model. For this reason, federated learning is often coupled with database privacy preservation techniques for preventing attacks, such as the addition of differentially private noise \citep{dwork2006differential}. Strong encryption schemes are also commonly employed in the federated learning pipeline.

Federated learning has been researched specifically in computer vision settings as well. \cite{liu2020fedvision} created Fedvision, a platform to specifically support the use of federated learning in computer vision settings, such as for object detection. It has also prominently been used in medical imaging settings, medical images being highly privacy-sensitive in nature \citep{yan2020variation, silva2019federated}.

\subsection{System Level}
For system level privacy preservation, techniques need to be developed so that the data used in the pipeline becomes secure, and that user consent for the use of the data in the pipeline is traceable. Traceability requires two components \citep{mihaildis2020methodological}. The first is that personal data can be traced to when user consent for its usage was recorded. Secondly, the flow of the data to various sources should also be traceable. This is essential because withdrawal of consent is an important facet of privacy laws like the GDPR \citep{gdpr2018}; upon withdrawal of consent, actions have to be taken by the authorised administrator to comply with the request. For this reason, system level privacy is not only an essential concept, but also an arguably overlooked one that is critical to managing the legal requirements surrounding the use of data in machine learning projects. 

Additionally, an important facet to system level privacy is the creation of secure databases that protect against information breaches. State-of-the-art techniques like homomorphic encryption allow for machine learning models to infer from the data privately. \cite{boulemtafes2020review} provide a more in-depth treatment on the subject of privacy preserving deep learning. Techniques under secure processing (see Section \ref{subsec:secureproc}) can be considered as contributing to the system level privacy in a  system that enforces privacy by design, as for system level privacy, it is required that the data remains secure inside the pipeline. Secure processing techniques assist the pipeline in this regard. It is, however, unclear whether techniques categorised as secure processing also fall under model level privacy preservation schemes as they do allow models to infer information from the data, while also preserving user privacy. 



\subsection{User Interface Level}


Privacy provided at the user interface level, prevents the exposure of privacy-sensitive images or parts of images in various scenarios. Under the classification of privacy preservation methods proposed in this review, techniques under the category of visual obfuscation (Section \ref{sec:vis_obfuscation}) can be mentioned as adding to user interface level privacy of pipelines. Data hiding methods also contribute to the user interface level privacy of a pipeline because, according to definition, these act to restrict the exposure of private visual information within the image, differing from the former category by the strategy with which the hiding of sensitive information is performed. 

\subsection{User Level}
User level privacy measures empower users by helping them manage their data. These also help users understand the privacy risks involved with the sharing of their data, and also give them mechanisms through which they can control the disclosure of their data. User level privacy is ensured through various educative measures, such as through the use of clear and easy to understand privacy disclosures and agreements. The creation of transparent dashboards through which users can control their data usage is another measure. The regular collection, analysis, and incorporation of user feedback into the pipeline is also a measure to incorporate user level privacy into the pipeline.

\section{Performance Evaluation}
\label{sec:perfeval}
For the case of visual obfuscation techniques, the type of performance evaluation used depends on the adversary. In systems to perform machine obfuscation, image quality metrics \citep{pedersen2012full} are popularly used. Since the objective of machine obfuscation techniques is to create images that are perceptually similar to the original, image quality metrics are employed to ascertain the (dis)-similarity of the two images. As for perceptual obfuscation, where the adversary is a human observer, a more empirical evaluation is often used. Human feedback is commonly sought for this purpose through the deployment of targeted surveys. Machine recognition systems are also often employed in the case of facial de-identification tasks.

The following subsections deal with the most commonly used metrics in the literature. Popular datasets used during evaluation are also explained.

\subsection{Technical Privacy Metrics}
There are different types of privacy metrics that have been employed for measuring the performance of privacy preservation methods. \cite{wagner2018technical} refer to 8 categories of metrics used to measure privacy in various contexts. We classify technical privacy metrics into two strains: those which measure an adversary's estimates to gauge how private a dataset is, and those metrics which gauge privacy according to a variable independent of adversarial estimates. 


\subsubsection{Indistinguishability Metrics}
Indistinguishability metrics measure whether an adversary can distinguish between two outcomes of a \textit{privacy mechanism}, and gather information about the dataset's composition from the differences between the outcomes. One commonly used indistinguishability metric is differential privacy \citep{dwork2014algorithmic}, which is nowadays extensively used in the securing of databases.

\cite{dwork2014algorithmic} define differential privacy as a promise made by a data holder/curator to a data subject. The promise is defined as follows:
\begin{quote}
    You will not be affected, adversely or otherwise, by allowing your data to be used in any study or analysis, regardless of what other studies, datasets or information sources are available.  
\end{quote}

When differential privacy is implemented for a specific database, it ensures protection against differencing attacks that can reveal information about a specific user in the database. By assuring differential privacy, the designer is ensuring that upon removal of a record containing a specific user's information, queries executed against the database do not produce a different output from when the same query was executed on the version of the database with the user's record present.   

In obfuscation tasks, a commonly used metric is the accuracy of machine recognition systems, which \cite{wagner2018technical} classify into error-based metrics. This looks at how often a machine recognition system like Amazon Rekognition engine \citep{amazonrekognition} can identify subjects in images that have been visually obfuscated. It is usually the case that a simple tally is used as the metric, counting the number of times the subject of interest is detected.

Of particular interest to the concept of perceptual obfuscation are metrics that are independent of adversary. These are solely dependent on observable or measurable differences between two data points or sets of data.

\subsubsection{Data Similarity Metrics}
One such category proposed by \cite{wagner2018technical} is data similarity. These include metrics that measure the similarity within a dataset through the formation of equivalence classes, or between two sets of data. Some common types include $k$-anonymity \citep{doi:10.1142/S0218488502001648} and its variants, namely $l$-diversity \citep{10.1145/1217299.1217302} and $t$-closeness \citep{li2007t}.  
\\~\\
\textbf{$k$-Anonymity}~— $k$-anonymity is one of the most widely used metrics to evaluate privacy and defines itself regarding quasi-identifiers inside a database. \textit{Quasi-identifiers} are attributes that can be taken together to identify an individual. Examples of this include the postcode or the birthdate in a personal database. In the case of a facial features database, this can refer to features like glasses, shapes of facial features like noses and the face itself. The metric is defined as follows - 

\begin{quote}
    A database is private if each record, $k$, in the database is indistinguishable from at least $k-1$ records in the database with quasi-identifiers.
\end{quote}
 Upon satisfaction of $k$-anonymity, a person's record can only be chosen from a database with a probability of $1/k$.
\\~\\
\textbf{$l$-Diversity}~— Proposed to address the limitations of $k$-anonymity, $l$-diversity is defined as follows - 

\begin{quote}
    For the equivalence class representing a set of records with the same values for quasi-identifiers, it should have at least $l$ `well-represented' values for the sensitive attribute.
\end{quote}

`Well represented' values are commonly defined as whether an equivalence class has $l$ distinct values for the sensitive attribute, without considering the frequency of values. A stronger version of the metric is called \textit{entropy $l$-diversity}, defined as follows

\begin{align*}
     \textnormal{Entropy}(E) &\geq \textnormal{log}\: l \\
    \textnormal{Entropy}(E) &= -\sum_{s \in S}p(E, s)\;\textnormal{log}\: p(E, s)
\end{align*}
 
\noindent with $E$ being the equivalence class, $S$ being the value set of the sensitive attribute, and $p(E, s)$ is the fraction of records in $E$ that have sensitive value $s$.
\\~\\
\textbf{$t$-Closeness} - To prevent attacks on privacy by adversaries with knowledge of global distribution of sensitive attributes inside a database, \cite{li2007t} devised the measure of $t$-closeness. This measure updates $k$-anonymity as follows.

\begin{quote}
    The distribution of sensitive values, $S_E$, in an equivalence class $E$ shall be close to its distribution, $S$ inside the entire database. 
    \begin{align}
        \forall E : d(S, S_E) \leq t 
        \label{eq:tclose}
    \end{align}
    with $d(S, S_E)$ being the distance between distributions $S$ and $S_E$ measured by the Earth mover distance \citep{andoni2008earth}, and $t$ is a privacy threshold that should not be exceeded.
\end{quote}

\subsubsection{Machine Recognition Scores}
Particularly in the context of facial de-identification, machine recognition is commonly employed as a metric to gauge the effectiveness of obfuscation methods. Machine recognition algorithms work by scoring how often a trained recognition algorithm can identify a de-identified subject. In the context of facial recognition, the most commonly used API services are the Google Vision API \citep{googlecloudvision}, Microsoft Azure Face API \citep{azurefacial}, Amazon Rekognition \citep{amazonrekognition} and Face++ \citep{faceplusplus}. Simple scoring systems are mostly used for these metrics, often a simple tally of the recognised attribute in the case of attribute recognition, or the recognised activity category in the case of an activity recognition task on obfuscated frames.

For gait obfuscation, custom metrics are usually employed. \cite{tieu2019rgb} craft custom automatic evaluation strategies that seek to measure the difference between a standard gait and a generated one. They employ a \textit{frame score} and a \textit{video score} to measure the differences. The \textit{frame score} measures the degree to which the shape of the object in the frame looks human. For this, they employed a pretrained YOLO model \citep{redmon2018yolov3} that detects and classifies objects in an image. The authors compute the probability that a person in a frame belongs to the `person' class. The \textit{video score} measures the degree to which the gait in the video looks like a humanoid walking. A pretrained ResNeXT-101 \cite{xie2017aggregated} was used for this purpose, which classifies actions in the video. The probability that the action in the video corresponds to the `video' class is measured and reported for this score. 

\subsubsection{Human Recognition Scores}
Humans are also often employed to gather user feedback about the efficacy of privacy preservation methods. Targeted questionnaires are often employed for this purpose. These are mostly composed of questions that allow the respondents to select from an available set of options. Some questionnaires featuring a free form-filling section where the respondents can fill in their responses. As an alternative to in-person surveys, various services have also sprung up over the years that helps researchers solicit responses from targeted audiences. Arguably the most popular ones seen in literature are the Mechanical Turk \citep{amazonmechanicalturk} and Prolific \citep{prolific}. 

\cite{cciftcci2015using} created an interface through which the authors solicit responses from participants who are tasked with recognizing faces after image filtering (using the `false colors' method) is done. They also asked participants to do activity recognition on filtered videos. \cite{padillalopez2015visual} employed targeted questionnaires to quantify the degree of privacy provided by various perceptual privacy preservation methods. The methods that were chosen were, in no order: blurring, pixelation, embossing, silhouette, skeleton, and an avatar. Visuals after applying these methods were shown to participants. The participants were then tasked to answer questions that delved into perceptual attributes such as the colour of the obfuscated subject's hair, skin colour, and whether subjects are smiling in the frames presented.  

\subsection{User Acceptance Studies}
Another important concept is the acceptance of the privacy preservation technology in use. The work done by \cite{wilkowska2021video} is one such example, where the authors create a user survey to understand preferences of German and Turkish participants of lifelogging studies. The study examines and compares perspectives of users from the two countries to understand whether cultural influences affect perceptions of lifelogging technologies and the visual obfuscation techniques that are commonly used on these feeds. As part of the study, the researchers created visualisations of various privacy preservation techniques, selecting representative images that were obfuscated in 5 different ways. In order from low to high levels of privacy protection, these are: real image, Pixelation, Solid Silhouette, Avatar, and Skeleton. They invited a diverse set of users to then provide feedback on the images to answer, among others, the following question: ``Do German and Turkish participants choose similar or different visualisation modes of information representation for their use, be it in the context of the accepted location or regarding the data access for the others?" It also addresses the question of which visualisation mode being the most preferred one for the participants of the survey.

\subsection{Datasets}
The research community has employed several datasets for the task of measuring visual privacy. The most commonly used datasets consist of RGB images or video streams. It is also popular to curate subsets of these datasets for various targetted experiments. In this section, various datasets that are used for validating the efficacy of privacy preservation methods are listed, along with details of their composition and the papers that use these sets for experimentation\footnote{It is to be noted that a number of these datasets presented are aimed at measuring the efficacy of machine obfuscation methods}.


For the case of facial anonymisation, some popular datasets used are the following: 

\textbf{Facial Recognition Technology (FERET)} dataset \citep{phillips1998feret}~— 
Containing 14,126 facial stills of 1,199 people, FERET is a publicly available dataset from the US Army. For every facial image, the coordinates for the centres of the eyes and tip of the nose are provided. Examples of privacy preservation methods using FERET for validation include \cite{cciftcci2015using}.

\textbf{People in photo albums (PIPA)} dataset \citep{zhang2015beyond}~— is a dataset consisting of over 6,000 images of around 2,000 persons, with only half of the images being of persons from a frontal frame of reference. This creates a challenging task, as recognition systems are mostly trained on frontal imagery. The dataset contains people in a good variety of poses, activities, and scenery. One example of a method validated using PIPA is the method proposed by \cite{sun2018natural}.

\textbf{AT\&T Database of Faces} \citep{ATT}~—The AT\&T database of faces contains 400 grayscale images of 40 individuals of resolution 92$\times$112. The dataset contains 10 images of each individual, taken under a variety of conditions including varied lighting, different expressions, and different facial details. One example privacy protection scheme that uses this dataset for testing is that by \cite{fan2018image}.

\textbf{Facescrub} \citep{ng2014data} is a large dataset consisting of slightly more than 65,000 facial images of 530 celebrities collecting from online publications. Only URLs are distributed for copyright reasons\footnote{The original dataset contains URLs to 100000 images, with a number of URLs broken due to missing media.}. \cite{shan2020fawkes} proposes a scheme that makes use of this dataset while testing. 

\textbf{PubFig images dataset} \citep{kumar2009attribute}~— This is a dataset of images of public figures (celebrities and politicians) obtained from the internet. The dataset consists of around 60,000 images, with around 300 images per individual. \cite{shan2020fawkes} and \cite{sharif2016accessorize} are notable examples of papers using the PubFig images dataset.

\textbf{CelebFaces Attributes} (CelebA) dataset \citep{liu2015deep}~— Used for facial attribute estimation in the process of training facial de-identification methods, this dataset contains 202,599 images and 10,177 identities of celebrities. Each image has around 40 boolean attribute labels. \cite{li2019anonymousnet} is notable for making use of the CelebA dataset for testing.

\textbf{Labeled Faces in the Wild (LFW)} dataset \citep{huang2008labeled} is another dataset containing $\approx$13,000 images of faces collected from the web. 1,680 individuals in the set have two or more distinct images of themselves represented in the dataset. Several alternative datasets of faces in the wild have also been proposed, some notable ones being \textit{Fine-grained LFW} \citep{deng2017fine}, \textit{LFWGender} \citep{jalal2016lfw}, and \textit{LFW3D}. \cite{zhang2021multi} proposes a method that is notable for using the LFW dataset during testing. 

Generic image recognition and object detection datasets are often used in validating the efficacy of privacy preservation schemes, mostly in the case of machine obfuscation schemes. Some commonly used ones are the following. 

\textbf{Modified NIST (MNIST)} \citep{lecun1998mnist}~— MNIST is an extremely popular dataset consisting of images of handwritten digits collected from census bureau employees and high school students in the USA. The entire dataset consists of 70,000 images in total. \cite{abadi2016deep} proposes a scheme that is benchmarked using the MNIST dataset.

\textbf{CIFAR-10} \citep{krizhevsky2009learning}~— Another popular dataset is CIFAR-10, consisting of  of a total of 60,000 images of size 32$\times$32. Labels of the dataset consists of either animals (e.g., cats, dogs etc.), or vehicles (e.g., planes, cars, etc.). \cite{abadi2016deep} proposes a scheme that uses the CIFAR-10 dataset for validation.

\textbf{YouTube 8M video dataset} \citep{abu2016youtube}~— The YouTube 8M dataset is a video dataset composed of around 8million videos, approximately 500,000 hours of content, annotated in a multi-label format with 4,800 distinct labels.  These labels are machine generated and human curated, with 1.9 billion video frame-level annotations. The entities in videos are also categorised, with some categories represented in the dataset being 'Arts \& Entertainment', 'Games', 'People \& Society', and 'Books \& Literature'. \cite{wong2020privacy} proposed a privacy preservation scheme that notably uses the YouTube 8M video dataset for testing. 


In the setting of gait anonymisation, the \textbf{CASIA-B gait dataset} \citep{yu2006framework} is one that is arguably most popular. This dataset contains 124 individuals in total, with 110 sequences (10 sequences each for each of 11 viewing angles from $0^{\circ}$ to $180^{\circ}$). \cite{tieu2017approach} create a gait anonymisation scheme that uses the CASIA-B dataset for validation.

In the context of full-body de-identification, the following datasets are commonly used~— 

\textbf{Clothing Co-Parsing dataset} \citep{yang2014clothing}~— This dataset consists of 2,098 high resolution, street fashion images. Pixel-level segmentations of individual garments and skin are available for $\approx$1000 of the images. 59 segmentation tags defining various garment types, e.g., blazer, cardigan, sweatshirt etc., are used in this dataset. \cite{brkic2017know} makes use of the clothing co-parsing dataset to test their full-body privacy preservation scheme. 

\textbf{Human3.6M dataset} \citep{ionescu2013human3}~— This dataset consists of 3.6 million video frames of actors performing actions in a controlled setting. 3D joint positions, the laser scans of the actors, and their corresponding 3D poses are available as annotations. The dataset utilises a static camera angle for the recordings. \cite{brkic2017protecting} proposed a privacy protection scheme that utilised this dataset for testing purposes. 

\textbf{Toyota Smarthomes dataset} \citep{das2019toyota}~— This is a dataset of slightly more than  16,000 video clips, of 31 activity classes performed by 18 seniors in a smarthome setting. The dataset is labelled with both coarse and fine-grained labels and contains heavy class imbalances, high intra-class variation, simple as well as composite activities, and activities with similar motion and of variable duration. \cite{climent2021protection} use the Toyota Smarthomes dataset to validate their privacy preservation scheme.

\textbf{NTU RGB+D dataset} \citep{Shahroudy_2016_CVPR}~- Containing 60 different action classes including daily, interaction-based, and health-related actions, this is a large-scale dataset for RGB+D human action recognition, containing greater than 56,000 samples and 4,000,000 frames, collected from 40 distinct subjects. \cite{Wang_2019_CVPR_Workshops} use this dataset to test the efficacy of their privacy preserving action recognition method. An extended version of this dataset was published by \cite{8713892}.

\section{Conclusion and Future Directions}
\label{sec:conclusion}
This work reviews the state of the art in visual privacy preservation methods. A low-level taxonomy of visual privacy preservation methods is introduced, and the categories under the taxonomies were subsequently explored. Special attention was given to visual obfuscation methods, these being of most relevance to AAL applications. The taxonomy is then connected to a high-level classification scheme of the levels of privacy by design.

Visual obfuscation methods are categorised into two categories in this review based on the targets from whom the algorithms are seeking to hide private information: \textit{perceptual obfuscation} and  \textit{machine obfuscation} methods. Perceptual obfuscation seeks to perceptually alter images in ways that unauthorised human observers who view the visual feed are thwarted. By contrast, machine obfuscation methods try to hide privacy-sensitive elements from machine learning algorithms. These seek to alter the feature space of images in ways that machine recognition systems are thwarted, while also perceptually changing the visuals to the least possible extent. 

As these are two different directions of research, algorithms can also be built such that they perform both machine and perceptual obfuscation. The capability of performing reversible transformations through secure pipelines is another promising direction for research. This is useful in the case when reversibility is required, such as for an arbiter (a judge, a doctor, etc.) to view unedited footage to obtain full information about a specific scenario. 

\subsection{Technical Questions}
In the context of visual privacy preservation, numerous technical challenges remain to be addressed. One major challenge is to create real-time pipelines that impart privacy. Most of the existing state-of-the-art methods rely on computationally intensive pipelines. To create real-time privacy protection, methods have to be made more lightweight.

There are also some widely used cameras that are arguable not sufficiently researched in literature from the perspective of privacy preservation. Egocentric/wearable cameras have been touted as a method to protect identity, but this poses problems if the environment contains objects (e.g. mirrors) that reveals one's personal attributes. This also introduces issues when bystanders come into the visual field; bystanders would typically not have given permissions for them to be captured on camera. This poses ethical and legal challenges, in addition to technical ones when egocentric cameras are utilised~\citep{gurrin2014privacy}.

Omnidirectional cameras have fisheye lenses that provide the user with a mostly non-occluded view of an entire room based on its placement (usually on the ceiling). However, object detection algorithms have not typically been trained to detect on images from distorted lenses. Privacy preservation algorithms that rely on detection as part of the pipeline are therefore summarily excluded from use on these streams. Other non-standard cameras (thermal, infrared) also face similar problems. Therefore, the authors call for more research to create privacy preserving algorithms that work on non-standard cameras.

Some identifiers have also been arguably addressed less in the literature. Gait is one such example, and to the authors' knowledge, only a few papers have attempted to create gait anonymisation algorithms. Environmental identifiers are also another.

\subsubsection{Environmental Privacy}
\label{subsec:envpriv}

Although included in this review as a sub-category of perceptual obfuscation, literature searches show that environmental privacy is an under-researched area, but arguably one that is critical to the ensuring of visual privacy. Most of the existing methods that impart privacy target people and their visible attributes. However, objects in the environment are also required to be obfuscated if the identity of the person is to be protected. Objects like credit cards and address labels create privacy risks if not obfuscated. Some methods do, however, provide environmental privacy as a side effect upon their use. As an example, consider a blurring filter. When a blurring filter is applied to an image as a whole, textural information is lost, which might lead to smaller privacy-sensitive objects such as credit cards (and specifically the numbers printed on them) being obfuscated. Depending on the parameters used for the blurring, however, larger objects in the environment might still contribute to privacy leakages. 

\subsection{Social Scientific and Legal Aspects of Privacy}
There is also the urgent need to understand the methods from social scientific and legal perspectives. There needs to be studies to ascertain the level of acceptance of different perceptual obfuscation methods among the monitored subjects. It is also unclear as to the extent of the acceptability of reversible transformations for the subjects being monitored. Although there are several methods that reconstruct obfuscated images, the acceptability of reconstructed images through a reverse transformation pipeline that contains embedded stochasticity is an especially interesting one to study. In a setting such as that of a court or in forensics, as reconstruction is an imperfect process, there is always the possibility of information loss. It is unclear if such images are viable for presentation in such circumstances. There also needs to be more studies that detail the relationship between human perception and the metrics that are used to measure perceptual obfuscation. Although there are some studies that do this, there is a distinct need for more wide-ranging targeted studies to be performed. 

The concept of a `privacy paradox' also needs to be investigated. It is a known phenomenon that people act in contrast to what they believe their privacy preferences are, especially when it comes to their online behaviour~\citep{BARTH20171038}. Users claim to be concerned about their online privacy, but they do little to protect their personal data. If this is also the case for visual data like that used in AAL applications, then the gathering of subjective data about user preferences through a medium such as questionnaires should be called into question. It could mean that better ways of gauging preferences should be created and deployed. It could also mean that existing studies that gauge privacy preferences ought to be re-evaluated.

\section*{Data Availability}
Data sharing not applicable to this article as no datasets were generated or analysed during the current study.

\section*{Acknowledgements} 
This work is part of the visuAAL project on Privacy-Aware and Acceptable Video-Based Technologies and Services for Active and Assisted Living~(\url{https://www.visuaal-itn.eu/}). This project has received funding from the European Union’s Horizon 2020 research and innovation programme under the Marie Skłodowska-Curie grant agreement No 861091. The authors would also like to acknowledge the contribution of COST Action CA19121 - GoodBrother, Network on Privacy-Aware Audio- and Video-Based Applications for Active and Assisted Living~(\url{https://goodbrother.eu/}), supported by COST (European Cooperation in Science and Technology)~(\url{https://www.cost.eu/}).

\bibliographystyle{unsrtnat}
\bibliography{references}  






\end{document}